\documentclass[11pt]{article}

\usepackage[final]{acl}
\usepackage{times}
\usepackage{latexsym}
\usepackage{xspace}
\usepackage[T1]{fontenc}
\usepackage[utf8]{inputenc}
\usepackage{microtype}
\usepackage{inconsolata}

\usepackage{amsmath,amssymb}
\usepackage{booktabs}
\usepackage{multirow}
\usepackage{tabularx}
\usepackage{array}
\usepackage[table]{xcolor}  

\usepackage{graphicx}

\usepackage{CJKutf8}
\newenvironment{zh}{\begin{CJK*}{UTF8}{gbsn}}{\end{CJK*}}

\usepackage{hyperref}
\usepackage{enumitem}
\setlist[itemize]{leftmargin=*,itemsep=0.5pt,topsep=2pt}
\setlist[enumerate]{leftmargin=*,itemsep=0.5pt,topsep=2pt}

\title{\textsc{CArtBench}: Evaluating Vision-Language Models on Chinese Art Understanding, Interpretation, and Authenticity}

\author{
  Xuefeng Wei$^{1}$\quad
  Wang Zhixuan$^{2}$\quad
  Xuan Zhou$^{1}$\quad
  Zhi Qu$^{1}$\quad
  Hongyao Li$^{2}$\\
  \textbf{Yusuke Sakai$^{1}$\quad
  Hidetaka Kamigaito$^{1}$\quad
  Taro Watanabe$^{1}$}\\
  $^{1}$Nara Institute of Science and Technology\quad
  $^{2}$Liaoning Normal University\\
  \texttt{\{xuefeng.wei.yb1, xuan.zhou.xc7\}@naist.ac.jp}\\
  \texttt{\{qu.zhi.pv5, sakai.yusuke.sr9, kamigaito.h, taro\}@is.naist.jp}\\
  \texttt{\{wang, lihongyao1989\}@lnnu.edu.cn}
}

\newcommand{\best}[1]{\cellcolor{pink!40}\textbf{#1}}

\newcommand{\benchmark}{\textsc{CArtBench\xspace}}
\newcommand{\curatorqa}{\textsc{CuratorQA\xspace}}
\newcommand{\catalogcaption}{\textsc{CatalogCaption\xspace}}
\newcommand{\reinterpret}{\textsc{ReInterpret\xspace}}
\newcommand{\connoisseurpairs}{\textsc{ConnoisseurPairs\xspace}}

\begin{document}
\maketitle

\begin{abstract}
We introduce \benchmark, a museum-grounded benchmark for evaluating vision-language models (VLMs) on Chinese artworks beyond short-form recognition and QA. \benchmark comprises four subtasks: \curatorqa{} for evidence-grounded recognition and reasoning, \catalogcaption{} for structured four-section expert-style appreciation, \reinterpret{} for defensible reinterpretation with expert ratings, and \connoisseurpairs{} for diagnostic authenticity discrimination under visually similar confounds. \benchmark{} is built by aligning image-bearing Palace Museum objects from Wikidata with authoritative catalog pages, spanning five art categories across multiple dynasties. Across nine representative VLMs, we find that high overall \curatorqa{} accuracy can mask sharp drops on hard evidence linking and style-to-period inference; long-form appreciation remains far from expert references; and authenticity-oriented diagnostic discrimination stays near chance, underscoring the difficulty of connoisseur-level reasoning for current models.
\end{abstract}

\section{Introduction}
Vision-Language Models (VLMs) are increasingly deployed as general-purpose multimodal assistants, yet their evaluation remains dominated by web images and Western-centric concepts \citep{geirhos2020shortcut,agrawal2018dontjustassume,ananthram2024seeit}.
Recent Chinese and culture-focused benchmarks broaden coverage but largely emphasize short-form recognition and QA \citep{nayak-etal-2024-benchmarking,cvlue2024,lan2025mcbe}.
As a result, a key capability remains under-tested: expert-facing interpretation that is culturally grounded and explicitly supported by visual evidence.
This matters in practice as VLMs are increasingly considered for cultural heritage access and documentation, where culturally misaligned or unsupported interpretations reduce reliability and downstream usability.

Chinese art makes this evaluation gap particularly salient.
Many visual conventions are period-sensitive, and curator-facing understanding often requires linking observable cues, such as motifs, brushwork, seals, material appearance, and format, to coarse historical context rather than relying on surface recognition alone.
We operationalize this requirement in our benchmark through knowledge-involved questions and style-to-period inference, and our experiments show consistent degradations on these subsets even for strong models (see Table~\ref{tab:curatorqa-main-breakdown}).
Authenticity judgment is another central workflow and often depends on subtle global consistency under visually similar confounds \citep{chen2025acpas}, yet we find that current VLMs remain near chance in this setting (Table~\ref{tab:task4_pairs}).
Together, these observations motivate a benchmark that measures not only short-form correctness, but also evidence linkage, structured appreciation, defensible interpretation, and confound-aware discrimination.

Concretely, \benchmark{} addresses four research questions:
\textbf{(RQ1)} Can VLMs answer museum-grounded questions with explicit evidence grounding?
\textbf{(RQ2)} Can they produce structured appreciation grounded in visible details?
\textbf{(RQ3)} Can they generate reinterpretations tethered to visual evidence?
\textbf{(RQ4)} Can they reason about authenticity cues without overconfident hallucination?

\begin{figure*}[!t]
  \centering
  \includegraphics[width=0.75\linewidth]{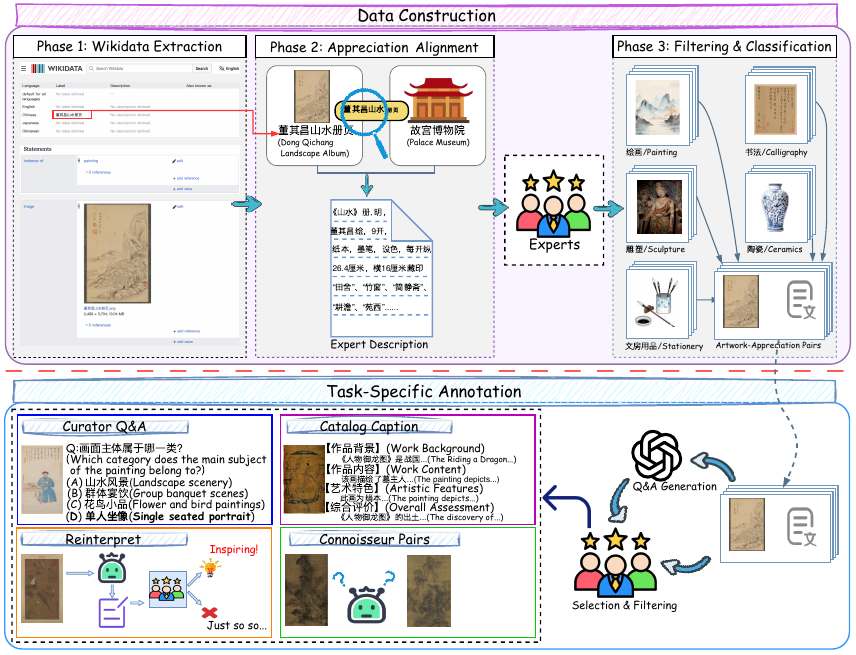}
  \caption{Overview of \benchmark\ construction and task instantiation. \textbf{Top: Phase 1} retrieves image-bearing Palace Museum objects from Wikidata, \textbf{Phase 2} aligns them to official catalog pages to collect curatorial descriptions, and \textbf{Phase 3} performs expert filtering and category assignment to yield museum-grounded artwork--appreciation pairs. {Bottom:} the curated pairs are instantiated into four tasks: \curatorqa (evidence grounding), \catalogcaption (structured appreciation), \reinterpret (expert-rated reinterpretation), and \connoisseurpairs (confound-aware authenticity diagnostics).}
  \label{fig:pipeline}
\end{figure*}

We introduce \benchmark, a museum-grounded benchmark that targets curator-facing Chinese artwork understanding in a unified suite (Figure~\ref{fig:pipeline}).
\benchmark{} is built by aligning image-bearing Palace Museum objects from Wikidata with authoritative catalog pages, followed by expert-guided curation and controlled task instantiation \citep{vrandecic2014wikidata,jin-etal-2023-museumqa}.
It comprises four tasks: (\romannumeral1) \curatorqa{} for evidence-grounded recognition and reasoning (14{,}421 questions over 1{,}589 artworks), (\romannumeral2) \catalogcaption{} for structured four-section appreciation (86 artworks), (\romannumeral3) \reinterpret{} for defensible reinterpretation with two-stage human ratings (25 canonical works), and (\romannumeral4) \connoisseurpairs{} as a diagnostic stress test for authenticity discrimination on visually similar pairs (10 pairs).
We note that \reinterpret{} and \connoisseurpairs{} are intentionally small-scale due to the high cost of expert curation, and are best understood as diagnostic evaluations rather than statistically definitive leaderboard comparisons.

While culturally grounded multimodal evaluation is an active research direction \citep{nayak-etal-2024-benchmarking,lan2025f2bench}, museum-grounded evaluation for Chinese artworks remains rare, especially for long-form, interpretation-heavy tasks beyond recognition and short-form QA.
Existing Chinese heritage datasets primarily target conservation or recognition tasks such as mural restoration \citep{xu2024muraldh}, manuscript character detection \citep{liu2025deepjiandu}, and calligraphy classification \citep{zhao2025mccd}, leaving a gap for expert-facing, evidence-referenced multimodal reasoning.

Across nine representative VLMs, we find consistent failure modes detailed in Section~6.
Our contributions are:
\begin{itemize}
  \item We construct \benchmark\footnote{Our benchmark, evaluation code, and derived annotations are available at \url{https://github.com/Big-Sid/CARTBENCH-Chinese-Artwork-Benchmark}.}, a museum-grounded benchmark for Chinese art spanning recognition, appreciation, reinterpretation, and authenticity diagnostics, covering multiple dynasties and five art categories.
  \item We provide unified evaluation protocols combining automatic metrics, format compliance checks, rubric-based expert scoring, and human rubrics that assess evidence linkage, structured long-form compliance, and interpretive defensibility.
  \item We benchmark representative VLMs and identify systematic failure modes: high \curatorqa{} accuracy can coexist with sharp degradations on style-to-period inference and evidence linking, which are obscured by short-form evaluation alone.
\end{itemize}

\section{Related Work}

\paragraph{Vision--language evaluation beyond general VQA.}
Large-scale VQA and captioning benchmarks established standardized evaluation pipelines,
but they primarily target generic recognition and short-form responses, offering limited coverage for
\emph{evidence-linked} and \emph{domain-grounded} interpretation \citep{antol2015vqa, lin2014mscoco}. Comprehensive VLM evaluations such as MME \citep{fu2025mme} and MM-Vet \citep{yu2023mm} further highlight the need for systematic testing beyond general VQA.
Moreover, recent analyses of VLMs highlight that strong fluency may coexist with weak visual grounding,
manifesting as hallucinated or unsupported claims \citep{li-etal-2023-evaluating}.
These observations motivate evaluations that explicitly stress faithfulness and justification, which are central to art appreciation.

\paragraph{Cultural grounding and fairness in VLM benchmarks.}
To probe cultural and geo-diverse understanding, recent benchmarks have begun to move beyond Western-dominant imagery and concepts.
CVLUE constructs Chinese culture-driven image and category selections and reports consistent gaps between Chinese and English counterparts,
revealing cultural knowledge deficits under distribution shift \citep{cvlue2024}.
CulturalVQA similarly evaluates geo-diverse cultural comprehension and finds region-dependent disparities across facets such as rituals and traditions \citep{nayak-etal-2024-benchmarking}. Relatedly, cross-lingual studies of artwork explanation suggest that multilingual generation can further amplify faithfulness and cultural nuance challenges \citep{ozaki2025towards}.
Beyond cultural knowledge, fairness-oriented benchmarks have highlighted systematic biases in LLM outputs across demographic and cultural dimensions.
F$^2$Bench introduces an open-ended fairness evaluation with factuality considerations, demonstrating that strong models can still produce culturally biased or factually misaligned outputs \citep{lan2025f2bench}.
McBE provides a multi-task Chinese bias evaluation benchmark and reveals that even high-performing LLMs exhibit nontrivial biases specific to Chinese cultural contexts \citep{lan2025mcbe}.
These findings reinforce the need for culture-specific evaluation, but existing benchmarks mostly focus on short-form QA and do not directly assess expert-facing, long-form interpretive structure
or connoisseur-level reasoning.

\paragraph{Art- and museum-centered vision--language resources.}
Art-specific datasets explore dimensions that go beyond object naming.
For example, ArtEmis pairs artworks with viewer emotions and grounded explanations, enabling affective captioning and interpretation-like language generation \citep{Achlioptas_2021_CVPR}, and ArtELingo-28 extends art-language benchmarks with large-scale multilingual captions across 28 languages \citep{mohamed2024no}.
Museum-centered QA resources, such as MuseumQA, further emphasize artifact knowledge and fine-grained factual questioning in curated museum settings \citep{jin-etal-2023-museumqa}.
In parallel, recent efforts build large-scale QA systems for Chinese painting and calligraphy by integrating structured resources and external documents \citep{wan2025chpcqa}.
These datasets are complementary to ours: they typically emphasize factual correctness or affective response,
whereas \benchmark{} targets a unified suite that jointly evaluates curatorial reasoning, structured expert-style appreciation,
defensible reinterpretation, and authenticity-oriented diagnostic discrimination.

\paragraph{Authenticity and computational assistance.}
Authentication has attracted increasing attention as a high-stakes cultural heritage application.
ACPAS proposes an expert-assistance system for authenticating ancient Chinese paintings via LLM-based agents,
and ArtEyer develops a multimodal authentication system augmented with contextual visualizations \citep{chen2025acpas, chao2024arteyer}.
While these systems demonstrate the feasibility of computational support for connoisseurship,
they are not designed as standardized VLM benchmarks with controlled confounds and consistent scoring protocols. 
This motivates benchmark-oriented diagnostic evaluation settings for authenticity discrimination under visually similar confounds.

\section{\benchmark: Data Construction}
Motivated by the gaps identified above, we construct \benchmark{} through a museum-grounded pipeline that aligns Wikidata entities with authoritative catalog pages, followed by expert-guided curation and controlled annotation.
Figure~\ref{fig:pipeline} overviews the museum-grounded construction pipeline used to build \benchmark.
From the resulting museum-grounded artwork--description pairs, four subtasks are instantiated to evaluate complementary aspects of Chinese artwork understanding, as outlined below.

\begin{figure}[t]
\includegraphics[width=\columnwidth]{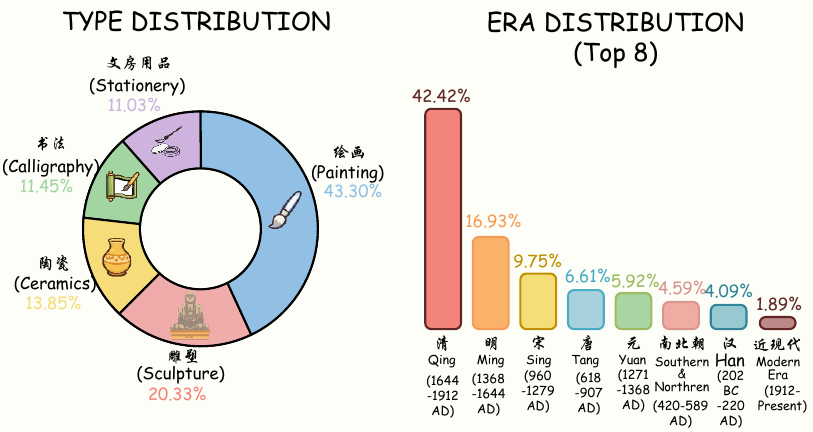}
\caption{Type and era distributions of \curatorqa\ entries: (left) five art categories; (right) top-8 merged eras.}
  \label{fig:3}
\end{figure}

\subsection{Alignment via Wikidata}
First, we retrieve museum objects with images using Wikidata Query Service \citep{vrandecic2014wikidata,guan2024hallusionbenchadvanceddiagnosticsuite}.
We restrict to objects associated with the Palace Museum (Beijing)\footnote{https://intl.dpm.org.cn/index.html?l=en} and require image availability.
This initial Wikidata query yields 127,601 image-bearing items linked to the Palace Museum.
We then align each object by title to the museum catalog page and collect the on-page curatorial description.

\subsection{Category and dynasty coverage}
Following expert feedback and Wikidata metadata, we assign each item to one of five categories: painting, calligraphy, sculpture, ceramics, and stationery.
The collection spans multiple dynasties to avoid overfitting to a single period.
Figure~\ref{fig:3} summarizes the resulting type and era distributions. For reference, Appendix~\ref{sec:appendix_dynasty_timeline} lists the dynasties and approximate date ranges used in our era bins.

\subsection{Task-specific annotation}
We build four tasks with complementary supervision and evaluation, targeting evidence-linked recognition (\curatorqa), structured appreciation (\catalogcaption), defensible reinterpretation (\reinterpret), and diagnostic authenticity discrimination (\connoisseurpairs).
The bottom of Figure~\ref{fig:pipeline} summarizes how the curated pairs are instantiated into four tasks.
We detail each task's format and annotation below; evaluation protocols are in Section~\ref{sec:evaluation}.

\paragraph{\curatorqa.}
\textbf{Task format.} Each instance provides an artwork image and a question; the model outputs one option (A/B/C/D) for multiple-choice items or \{True, False\} for verification.
This subtask aims to evaluate whether VLMs can answer curator-style questions with explicit visual evidence and calibrated domain knowledge requirements.
We generate question--answer pairs from aligned artwork images and authoritative museum catalog descriptions using GPT-5.2 \citep{openai_gpt52_model} under controlled prompting.
For multiple-choice items, GPT-5.2 jointly generates a set of plausible distractor options together with the correct answer during the same QA generation step.
GPT-5.2 also assigns the difficulty label (P1/P2) and the QA-type category following a fixed schema.

\textbf{Expert audit.}
To verify LLM-assisted generation quality, an art expert conducted a stratified random audit of 1{,}000 questions, stratified by difficulty (P1/P2) and QA type (QA1--QA6), with sampling capped at 3 per artwork (785 unique artworks).
The expert assessed gold-answer correctness, absence of ambiguity, and label consistency.
Only 1 erroneous item was found and corrected, yielding a 95\% upper-bound error rate of $\approx$0.47\% (exact binomial CI).
This low rate reflects multi-stage automated QC preceding the audit.

Prompt-based constraints and expert review remove unclear items and inconsistent question--answer pairs, yielding \textbf{14{,}421} questions over 1{,}589 artwork entries.
Each question is annotated with two difficulty settings: P1 is answerable from visible evidence alone, while P2 requires combining visual cues with art knowledge (e.g., inferring the dynasty from style).
Questions are categorized into six QA types, including subject recognition, scene/activity classification, composition or format, visible technique or style cues, iconography detection, and style-to-period inference.
Appendix~\ref{appendix:data-stats} reports the main statistics and distributions.


\begin{table*}[t]
\centering
\small
\setlength{\tabcolsep}{5.3pt}
\renewcommand{\arraystretch}{1.05}
\begin{tabular}{l c cccccc ccccc}
\toprule
\multirow{2}{*}{\textbf{Model}} & \multirow{2}{*}{\textbf{Acc.}} &
\multicolumn{6}{c}{\textbf{QA type Acc.}} &
\multicolumn{5}{c}{\textbf{Category Acc.}} \\
\cmidrule(lr){3-8}\cmidrule(lr){9-13}
& & QA1 & QA2 & QA3 & QA4 & QA5 & QA6
& Callig. & Station. & Paint. & Ceram. & Sculpt. \\
\midrule

\multicolumn{13}{@{}l}{\textbf{Qwen family}}\\
\hspace{1em}Qwen3-VL-235B & \best{0.84} & \best{0.88} & \best{0.94} & \best{0.88} & \best{0.89} & \best{0.56} & \best{0.81}
& \best{0.85} & \best{0.78} & \best{0.87} & \best{0.79} & \best{0.84} \\
\hspace{1em}Qwen3-VL-30B   & 0.80 & 0.85 & 0.91 & 0.85 & 0.88 & 0.42 & 0.78
& 0.83 & 0.74 & 0.83 & 0.76 & 0.80 \\
\hspace{1em}Qwen2.5-VL-72B     & 0.81 & 0.85 & 0.93 & 0.85 & 0.87 & 0.53 & 0.76
& 0.83 & 0.74 & 0.85 & 0.75 & 0.80 \\
\hspace{1em}Qwen2.5-VL-32B     & 0.80 & 0.84 & 0.91 & 0.83 & 0.87 & 0.53 & 0.74
& 0.82 & 0.76 & 0.83 & 0.73 & 0.78 \\
\hspace{1em}Qwen2.5-VL-7B      & 0.76 & 0.82 & 0.89 & 0.78 & 0.81 & 0.44 & 0.72
& 0.75 & 0.70 & 0.80 & 0.68 & 0.75 \\

\addlinespace[2pt]
\multicolumn{13}{@{}l}{\textbf{GLM family}}\\
\hspace{1em}GLM-4.5V & 0.57 & 0.59 & 0.67 & 0.62 & 0.62 & 0.29 & 0.56
& 0.59 & 0.52 & 0.60 & 0.54 & 0.57 \\
\addlinespace[2pt]
\multicolumn{13}{@{}l}{\textbf{GPT family}}\\
\hspace{1em}gpt-5-mini        & 0.79 & 0.84 & 0.90 & 0.83 & 0.86 & 0.46 & 0.78
& 0.83 & 0.74 & 0.84 & 0.72 & 0.78 \\
\hspace{1em}gpt-5-nano        & 0.59 & 0.62 & 0.73 & 0.53 & 0.65 & 0.27 & 0.61
& 0.59 & 0.55 & 0.63 & 0.53 & 0.56 \\

\addlinespace[2pt]
\multicolumn{13}{@{}l}{\textbf{Gemini family}}\\
\hspace{1em}gemini-2.5-flash  & 0.48 & 0.51 & 0.54 & 0.50 & 0.52 & 0.31 & 0.47
& 0.48 & 0.45 & 0.51 & 0.45 & 0.48 \\

\bottomrule
\end{tabular}
\caption{\curatorqa accuracy breakdown (overall, by question type, and by art category).
We report exact-match accuracy on 14{,}421 questions.
Category abbreviations: Callig.=calligraphy, Station.=stationery, Paint.=painting, Ceram.=ceramics, Sculpt.=sculpture.}

\label{tab:curatorqa-main-breakdown}
\end{table*}

\paragraph{\catalogcaption.}
\textbf{Task format.} Given an artwork image, the model generates an expert-style appreciation in four fixed sections:
Background ({\begin{zh}作品背景\end{zh}}), Content ({\begin{zh}作品内容\end{zh}}), Artistic Characteristics ({\begin{zh}艺术特色\end{zh}}), and Overall Evaluation ({\begin{zh}综合评价\end{zh}}).
We construct an expert-grounded set of 86 artworks via multi-stage curation: specialists collect candidates from professional Chinese art books and catalogues, we enforce Wikidata traceability to ensure each work has a stable image, and we filter low-quality images and non-informative texts.
Experts then \emph{rewrite} the reference appreciation into the same four-section schema, ensuring that references encode curator-grade insights and culturally appropriate terminology rather than raw scraped catalogue prose.
We release both a full reference and section-wise references for holistic and per-dimension evaluation.
This standardization reduces stylistic variance and enables automatic scoring and format compliance checks.

\paragraph{\reinterpret.}
\textbf{Task format.} Given an artwork image, the model produces an open-ended reinterpretation that goes beyond literal description while remaining image-grounded and culturally appropriate.
To enable a controlled assessment of interpretive novelty, art specialists selected \textbf{25} canonical Chinese artworks that are commonly used in formal art-training curricula and connoisseurship exercises.
These works have been extensively discussed in the field and typically come with established appreciation routines and conventional interpretations.
We therefore use them as anchors to test whether model interpretations can go beyond standard narratives while remaining grounded in the image and culturally appropriate.
Given the expert-curation cost, \reinterpret{} is best understood as a diagnostic evaluation rather than a basis for fine-grained model ranking.

\paragraph{\connoisseurpairs.}
\textbf{Task format.} Each instance presents an authentic--imitation pair; the model selects which artwork is authentic.
Imitations are common in Chinese painting due to its long history, and authenticity judgment requires cue-based connoisseurship beyond surface similarity.
Art specialists curated \textbf{10} pairs, each containing one genuine work and one visually similar replica, prioritizing close visual similarity, matched subject matter and medium, and stable provenance, while avoiding trivial cues such as obvious cropping artifacts.
\connoisseurpairs{} is positioned as a \emph{diagnostic stress test} rather than a basis for definitive model ranking, given the high cost of curating defensible pairs.

\section{Evaluation Protocols}
\label{sec:evaluation}

\subsection{\curatorqa}
\label{sec:eval_curatorqa}
We evaluate exact-match accuracy under constrained decoding.

\subsection{\catalogcaption}
\label{sec:task2_eval}
We evaluate \catalogcaption{} as a long-form, expert-style appreciation generation task with both automatic similarity metrics and structured format compliance.
To avoid confounds from Chinese word segmentation, we compute overlap-based metrics using \emph{character-level} tokenization, including ROUGE-L \citep{rouge}, BLEU-4 \citep{bleu}, and a CIDEr-like consensus similarity metric \citep{cider}.
We also report BERTScore (F1) \citep{bertscore} using \texttt{bert-base-chinese} with \texttt{rescale\_with\_baseline=True}.
As an auxiliary semantic signal beyond lexical overlap, we compute embedding cosine similarity (EmbSim) with \texttt{shibing624/text2vec-base-chinese}.
Because \catalogcaption{} targets a fixed four-section schema, we additionally measure \emph{Header rate}, the fraction of outputs that contain all four required section headers.
We note that these automatic metrics serve as scalable proxies for alignment to expert-curated references, and may not fully capture whether an appreciation is truly expert-like, culturally grounded, and well-supported by visual evidence. We therefore complement them with rubric-based expert evaluation (Section~\ref{sec:task2_results}).

Finally, we aggregate the core automatic metrics into a single KPI score:
\begin{equation}
\label{eq:kpi}
\begin{aligned}
\mathrm{KPI}
&= \lambda_{1}\,\mathrm{BERTScore}_{F1}
 + \lambda_{2}\,\mathrm{CIDEr\text{-}like} \\
&\quad + \lambda_{3}\,\mathrm{ROUGE}\text{-}\mathrm{L}
 + \lambda_{4}\,\mathrm{BLEU}\text{-}4,
\end{aligned}
\end{equation}
where $\lambda_{i} \ge 0$ and $\sum_{i=1}^{4}\lambda_i = 1$.
Unless otherwise specified, we use fixed weights $\{\lambda_{1},\lambda_{2},\lambda_{3},\lambda_{4}\}=\{0.45,\,0.25,\,0.20,\,0.10\}$.
We assign higher weights to semantic metrics (e.g., BERTScore) to better capture the interpretive nuance of appreciation text.
We further verify in Appendix~\ref{sec:appendix_task2_kpi_sensitivity} that our model rankings remain stable under diverse weight perturbations.

\subsection{\reinterpret}
\label{sec:eval_reinterpret}
We evaluate \reinterpret{} with a two-stage questionnaire adapted from the Torrance Tests of Creative Thinking (TTCT) \citep{torrance1974ttct}, redesigned for Chinese art appreciation.
Stage~1 performs a plausibility gate, filtering outputs with major misreadings, clear violations of widely accepted art-historical consensus, or unsupported factual fabrications.
Stage~2 rates the remaining outputs on a 1--5 Likert scale along five criteria:
\textbf{Interpretive Novelty} (D1), \textbf{Integrative Coherence} (D2), \textbf{Evidence-based Reasoning} (D3), \textbf{Elaboration \& Expressiveness} (D4), and \textbf{Perceived Creative Insight} (D5).
The full questionnaire is provided in Appendix~\ref{sec:appendix_task3_survey}.

\subsection{\connoisseurpairs}
\label{sec:eval_connoisseurpairs}
We evaluate \connoisseurpairs{} using pairwise accuracy over the authentic--imitation pairs.
To better understand failure mechanisms beyond a single accuracy number, we additionally conduct a qualitative error analysis with an expert-facing questionnaire; the full instrument is provided in Appendix~\ref{sec:appendix_task4_survey}.

\section{Experimental Setup}
We benchmark 9 representative vision--language models, including open-weight families
Qwen2.5-VL and Qwen3-VL \citep{qw25vl,qw3vl}, and GLM-V models \citep{glm45v},
as well as widely used API-based models from OpenAI (GPT-5 series) \citep{openai_gpt5,openai_gpt5mini,openai_gpt5nano,openai_gpt52}
and Google (Gemini 2.5 Flash) \citep{google_gemini25flash}.
We note that Qwen-family models are Chinese--English bilingual VLMs with strong Chinese multimodal performance on established benchmarks such as CMMU and CMMMU \citep{he2024cmmu,zhang2024cmmmu}, while GPT and Gemini models are general-purpose multilingual systems.
For each task, we use a unified prompt template and deterministic decoding, and we post-process outputs into
task-specific formats. Full prompting templates, decoding settings, and implementation details are in Appendix~\ref{sec:appendix_task1_prompt}--\ref{sec:appendix_llm_config}.

\paragraph{Human evaluation.}
Human evaluations for \reinterpret{} and \catalogcaption{} were conducted by three domain experts (two art graduate students and one art-school professor in China), all adult native Chinese speakers with formal training in Chinese art history. Annotators were compensated at rates meeting or exceeding local market standards; participation was voluntary with research-use notice, and we report only anonymized, aggregate results.

\section{Results and Analysis}
\subsection{\curatorqa}

\newcolumntype{Y}{>{\raggedright\arraybackslash}X}

\begin{table}[t]
\centering
\small
\setlength{\tabcolsep}{1pt}
\begin{tabular}{@{}l@{\hspace{20pt}} p{0.5\columnwidth} >{\raggedleft\arraybackslash}p{0.12\columnwidth}@{}}
\toprule
\textbf{Subset} & \textbf{mean $\Delta$ (CI$_{95}$)} & $\mathbf{p_{\text{perm}}}$ \\
\midrule
QA1 & 0.0426 [0.0279, 0.0574] & $<0.05$ \\
QA2 & 0.0354 [0.0215, 0.0485] & $<0.05$ \\
QA3 & 0.0435 [0.0301, 0.0568] & $<0.05$ \\
QA4 & 0.0345 [0.0224, 0.0462] & $<0.05$ \\
QA5 & \textbf{0.1022} [0.0735, 0.1302] & $<0.05$ \\
QA6 & 0.0295 [0.0169, 0.0423] & $<0.05$ \\
\addlinespace[2pt]
P1  & 0.0381 [0.0312, 0.0449] & $<0.05$ \\
P2  & \textbf{0.0626} [0.0488, 0.0766] & $<0.05$ \\
\addlinespace[2pt]
Calligraphy & 0.0254 [0.0061, 0.0435] & $<0.05$ \\
Painting    & 0.0294 [0.0199, 0.0385] & $<0.05$ \\
Sculpture   & 0.0678 [0.0522, 0.0827] & $<0.05$ \\
Ceramics    & 0.0688 [0.0508, 0.0883] & $<0.05$ \\
Stationery  & 0.0445 [0.0266, 0.0620] & $<0.05$ \\
\bottomrule
\end{tabular}
\caption{Subset-wise robustness of the performance gap between a Chinese--English bilingual proxy (Qwen3-235B) and a general-purpose API proxy (GPT-5-mini) on \curatorqa\.
We report $\Delta=\mathrm{Acc}(\text{Qwen3-235B})-\mathrm{Acc}(\text{GPT-5-mini})$ with 95\% confidence intervals (CI$_{95}$) obtained by bootstrap resampling.
Statistical significance is assessed using a sign-flip permutation test, with clustering at the artwork level to account for multiple questions per artwork.}

\label{tab:curatorqa-lang-strata}
\end{table}

\paragraph{Overall performance.}
Table~\ref{tab:curatorqa-main-breakdown} summarizes the main \curatorqa results on an aligned subset of 14{,}421 questions for which predictions can be consistently compared across model runs.
Among all evaluated models, the Qwen family dominates the leaderboard, with Qwen3-VL-235B achieving the best overall accuracy (0.84).
GPT-5-mini is the strongest general-purpose API baseline (0.79), while Gemini-2.5-flash is substantially lower (0.48).
Within the GLM family, GLM-4.5V attains moderate performance (0.57), but still trails the strongest Qwen models and GPT-5-mini.

\begin{table}[!t]
\centering
\small
\setlength{\tabcolsep}{3.2pt}
\renewcommand{\arraystretch}{1.04}
\begin{tabular}{@{}l r r r@{}}
\toprule
\textbf{Weights scenario} & \textbf{Spearman} & \textbf{Kendall} & \textbf{Top-1} \\
\midrule
Uniform weights                    & 0.991 & 0.964 & 1.000 \\
Dirichlet ($\alpha{=}5$)           & 0.991 & 0.964 & 1.000 \\
Dirichlet ($\alpha{=}1$)           & 0.964 & 0.891 & 1.000 \\
Local Gaussian ($\sigma{=}0.05$)   & 0.991 & 0.964 & 1.000 \\
\bottomrule
\end{tabular}
\caption{KPI weight sensitivity for \catalogcaption.
Spearman/Kendall are rank correlations with the default-weight ordering.
For stochastic schemes (Dirichlet and Local Gaussian), we report the median over samples.}
\label{tab:catalogcaption-kpi-robust}
\end{table}

\begin{table*}[t]
\centering
\small
\setlength{\tabcolsep}{4.2pt}
\begin{tabular}{l c c c c c c c}
\toprule
\textbf{Model} &
\textbf{KPI}$\uparrow$ &
\textbf{BERTScore$_{F1}$}$\uparrow$ &
\textbf{CIDEr-like}$\uparrow$ &
\textbf{ROUGE-L}$\uparrow$ &
\textbf{BLEU-4}$\uparrow$ &
\textbf{EmbSim}$\uparrow$ &
\textbf{Header rate}$\uparrow$ \\
\midrule

\multicolumn{8}{@{}l}{\textbf{Human baseline}} \\
\textbf{Human (mean of 3)} &
\textbf{0.698} & \textbf{0.746} & \textbf{0.673} & \textbf{0.699} & \textbf{0.546} & \textbf{0.982} & \textbf{0.988} \\
\midrule

\multicolumn{8}{@{}l}{\textbf{Qwen family}} \\
\hspace{1em}Qwen2.5-VL-72B   & \best{0.261} & \best{0.329} & 0.205 & \best{0.237} & \best{0.146} & 0.876 & 1.000 \\
\hspace{1em}Qwen3-VL-235B    & 0.258 & 0.319 & 0.223 & 0.223 & 0.138 & \best{0.893} & 1.000 \\
\hspace{1em}Qwen3-VL-30B     & 0.256 & 0.322 & \best{0.224} & 0.222 & 0.112 & 0.891 & 1.000 \\
\hspace{1em}Qwen2.5-VL-7B    & 0.245 & 0.305 & 0.196 & 0.224 & 0.137 & 0.873 & 1.000 \\
\hspace{1em}Qwen2.5-VL-32B   & 0.235 & 0.295 & 0.211 & 0.204 & 0.087 & 0.878 & 1.000 \\
\midrule

\multicolumn{8}{@{}l}{\textbf{GLM family}} \\
\hspace{1em}GLM-4.5V         & 0.245 & 0.315 & 0.194 & 0.217 & 0.113 & 0.878 & 0.988 \\
\hspace{1em}GLM-4.6V         & 0.241 & 0.302 & 0.209 & 0.211 & 0.100 & 0.882 & 0.977 \\
\midrule

\multicolumn{8}{@{}l}{\textbf{GPT family}} \\
\hspace{1em}gpt-5-nano       & 0.225 & 0.276 & 0.193 & 0.205 & 0.113 & 0.861 & 1.000 \\
\hspace{1em}gpt-5-mini       & 0.224 & 0.275 & 0.198 & 0.202 & 0.103 & 0.871 & 1.000 \\
\midrule

\multicolumn{8}{@{}l}{\textbf{Gemini family}} \\
\hspace{1em}gemini-2.5-flash & 0.170 & 0.201 & 0.160 & 0.160 & 0.069 & 0.812 & 0.256 \\
\bottomrule
\end{tabular}

\caption{\catalogcaption{} evaluation on 86 artworks.
Overlap-based metrics (ROUGE-L / BLEU-4 / CIDEr-like) use character-level tokenization.
BERTScore$_{F1}$ uses \texttt{bert-base-chinese} with \texttt{rescale\_with\_baseline=True}.
EmbSim is cosine similarity under \texttt{text2vec-base-chinese}. Header rate measures whether all four required
section headers appear in the output. References are expert-rewritten into the four-section schema.}
\label{tab:catalogcaption-main}
\end{table*}

\begin{table*}[!t]
\centering
\small
\setlength{\tabcolsep}{20.2pt}
\begin{tabular}{l c c c}
\toprule
\textbf{Model} &
\textbf{Gate pass}$\uparrow$ &
\textbf{Primary avg (D1--D3)}$\uparrow$ &
\textbf{All avg (D1--D5)}$\uparrow$ \\
\midrule
gemini-2.5-flash & 0.707 [0.600, 0.813] & 3.673 [3.476, 3.887] & 3.723 [3.528, 3.933] \\
GLM-4.6V         & 0.853 [0.760, 0.933] & 3.649 [3.522, 3.784] & 3.679 [3.575, 3.796] \\
gpt-5-mini       & 0.867 [0.787, 0.947] & 3.567 [3.480, 3.656] & 3.617 [3.529, 3.708] \\
qwen3-vl-235B    & \best{0.880} [0.787, 0.960] & \best{3.747} [3.622, 3.878] & \best{3.728} [3.621, 3.836] \\
\bottomrule
\end{tabular}
\caption{\reinterpret\ human creativity evaluation on 25 artworks (3 raters; 75 ratings per model).
Stage~1 is a plausibility gate; Gate pass is the fraction of outputs that enter Stage~2.
Stage~2 uses 1--5 Likert ratings and is computed only on passed outputs.
We report macro-averages over artworks with 95\% bootstrap confidence intervals.
Primary avg averages the three primary dimensions (D1--D3), and All avg averages all five dimensions (D1--D5).}
\label{tab:reinterpret-main}
\end{table*}

\paragraph{Category and QA-type effects.}

To substantiate subset-level difficulty patterns beyond aggregate accuracy, we test cross-model consistency by computing per-artwork accuracies and applying a sign test across models.
Category effects are systematic: for all models ($10/10$), \emph{stationery} and \emph{ceramics} are consistently harder than \emph{painting} (median gaps $-0.096$ and $-0.083$; $p{=}0.002$) and also harder than \emph{calligraphy} (median gaps $\approx -0.071$; $p{=}0.002$).
In contrast, QA-type variation is dominated by knowledge-heavy inference: QA5 (style$\rightarrow$period inference) shows the largest degradation across models (e.g., Qwen3-VL-235B: 0.56 on QA5 vs.\ $\ge$0.89 on QA1--QA4; GPT-5-mini: 0.46 vs.\ 0.85--0.91), aligning with QA5's intent to probe nontrivial stylistic-to-historical mapping.

\paragraph{Bilingual vs.\ general-purpose proxy gap.}
To assess whether models with strong Chinese multimodal capability exhibit a systematic advantage, we compare the best Chinese--English bilingual proxy (Qwen3-VL-235B) with the best general-purpose API proxy (GPT-5-mini).
As shown in Table~\ref{tab:curatorqa-lang-significance}, Qwen3-VL-235B improves by +4.51 points, significant under both paired McNemar and artwork-clustered tests.
Table~\ref{tab:curatorqa-lang-strata} confirms the advantage persists across question types and categories, with largest gains on knowledge-involved questions (P2) and QA5, which requires nontrivial style-to-period inference.
Our contribution here is not this observation per se, but that \benchmark{} makes such capability differences measurable and failure modes visible in a controlled setting.

\paragraph{Findings.}
Overall, \curatorqa{} yields high aggregate accuracy for strong models, yet exposes a consistent failure mode on style-to-period reasoning (RQ1).
The bilingual vs.\ general-purpose proxy comparison shows a statistically robust advantage for the bilingual proxy, with the largest gains concentrated in subsets requiring art-historical inference beyond direct visual recognition.

\subsection{\catalogcaption}
\label{sec:task2_results}
\paragraph{Results overview.}
Table~\ref{tab:catalogcaption-main} reports \catalogcaption{} results under the evaluation protocol in
Section~\ref{sec:task2_eval}.
Across all model families, automatic scores remain far below the human baseline, indicating that
expert-style appreciation generation is still a major bottleneck for current VLMs even when they can
follow the required format.

\paragraph{Cross-family comparison.}
The Qwen family consistently ranks highest on KPI, followed by GLM.
API baselines (gpt-5-mini/nano) are competitive but below the top Qwen variants, while \texttt{gemini-2.5-flash} performs substantially worse.
Strong models reach high embedding similarity (EmbSim $\approx 0.87$--$0.89$) yet low overlap metrics (ROUGE-L $\approx 0.20$--$0.24$), suggesting they capture coarse semantic intent but fail to match expert references in structured, section-specific detail.
We note that references are expert-rewritten into the four-section schema, not raw catalogue prose.

\paragraph{Complementary expert evaluation.}
Since overlap and embedding metrics may not fully capture genuine expert insight, we additionally conducted rubric-based expert evaluation with three raters on 258 model--artwork pairs (86 artworks $\times$ 3 models), scoring D1 (Evidence-grounded Interpretation), D2 (Cultural/Art-historical Appropriateness), D3 (Insight Depth) on 1--5 Likert scales, with moderate reliability (ICC(2,3) $\approx$ 0.62--0.64).
Results corroborate automatic metrics: both Qwen variants outscore Gemini on all dimensions, and the two Qwen variants are similar (details in Appendix~\ref{sec:appendix_task2_expert}).

\paragraph{Large gap to expert-written appreciation.}
Human writers achieve KPI $=0.698$ on average (three independent annotations), whereas the best model (Qwen2.5-VL-72B) reaches only $0.261$, attaining $\sim$37\% of the human KPI.
A similar trend holds across all component metrics: models substantially trail humans on BERTScore$_{F1}$, CIDEr-like, ROUGE-L, and BLEU-4, suggesting incomplete section-wise coverage and deviation from expert references in evidence-linked descriptions.

\paragraph{Robustness to KPI weighting.}
To ensure our conclusions do not hinge on a particular KPI weighting, we perturb metric weights using
uniform reweighting, Dirichlet sampling, and local Gaussian perturbations (Appendix~\ref{sec:appendix_task2_kpi_sensitivity}).
The top-ranked model is unchanged across all tested scenarios (Top-1 same $=1.0$), and rankings remain highly
consistent with the default ordering (Spearman 0.964--0.991; Kendall 0.891--0.964;
Table~\ref{tab:catalogcaption-kpi-robust}), supporting the robustness of the main findings.

\paragraph{Findings.}
\catalogcaption{} shows that expert-style, long-form appreciation remains far below human performance (RQ2).
The Qwen family consistently leads among evaluated models, yet both automatic metrics and expert scoring indicate weak reference alignment despite high semantic similarity.
The main conclusions are stable to KPI reweighting, and the complementary expert evaluation confirms that the gap is not an artifact of automatic metrics alone.

\subsection{\reinterpret}
\label{sec:task3_results}

\paragraph{Results overview.}
Table~\ref{tab:reinterpret-main} summarizes \reinterpret\ human ratings on 25 canonical artworks with three raters.
We report the Stage~1 plausibility gate pass rate and Stage~2 creativity scores on a 1--5 Likert scale.
Stage~2 scores are computed only for gate-passed outputs; all scores are macro-averaged over artworks with 95\% bootstrap CIs.

\paragraph{Inter-annotator agreement.}
To assess the reliability of human evaluation, we report IAA for both stages.
For Stage~1, Fleiss' $\kappa$ ranges from 0.415 to 0.711 across models (moderate to substantial agreement).
For Stage~2, ICC(2,3) ranges from 0.476 to 0.877 for Primary (D1--D3) and 0.649 to 0.888 for All (D1--D5), indicating that the averaged scores in Table~\ref{tab:reinterpret-main} are reasonably stable under rater variation.
Full per-model IAA statistics are in Appendix~\ref{sec:appendix_task3_iaa}.

\paragraph{Interpretation.}
Across models, Stage~2 scores on the passed subset fall into a relatively narrow range (Primary avg: 3.57--3.75).
This pattern is consistent with the two-stage design: Stage~1 filters clearly implausible outputs, so Stage~2 focuses on more subtle quality differences among responses that already meet a plausibility threshold.
Under our anchored rubric (Appendix~\ref{sec:appendix_task3_survey}), 3 corresponds to a competent, non-trivial reinterpretation,
whereas 4 reflects a clearly distinctive and well-supported reading.
In contrast, the Stage~1 gate provides a clearer separation: pass rates span 0.707--0.880,
suggesting that the dominant bottleneck is not stylistic expression but avoiding errors that invalidate an otherwise plausible interpretation.
Among gate failures, era/dynasty misattribution and key-content misreadings are the most frequent causes: for \texttt{gemini-2.5-flash}, 29.3\% of rater--artwork instances fail the gate, compared to 12.0\% for \texttt{qwen3-vl-235B}. Detailed failure-type distributions are in Appendix~\ref{sec:appendix_task3_stats}.

\paragraph{Findings.}
Overall, \reinterpret\ is constrained by defensibility (RQ3).
Models achieve higher effective creativity primarily by passing the plausibility gate more consistently, rather than by substantially
improving Stage~2 scores on already-valid outputs.
In this setting, \texttt{qwen3-vl-235B} is the most reliable, while \texttt{gemini-2.5-flash} is most limited by gate failures.
Given the diagnostic scale of this task, we caution against over-interpreting fine-grained score differences.

\subsection{\connoisseurpairs}
\label{sec:task4_results}

Table~\ref{tab:task4_pairs} reports pairwise authenticity discrimination on 10 authentic--imitation pairs.
Each example presents an A/B comparison, and a prediction is correct if the model selects the authentic work.

\begin{table}[!t]
\centering
\small
\setlength{\tabcolsep}{5pt}

\begin{tabular}{@{}l c@{\hspace{14pt}} l c@{}}
\toprule
\textbf{Model} & \textbf{Acc.} & \textbf{Model} & \textbf{Acc.} \\
\midrule
\multicolumn{4}{@{}l}{\textbf{GLM family}}\\
GLM-4.5V & \best{0.600} & GLM-4.6V & \best{0.600} \\
\midrule
\multicolumn{4}{@{}l}{\textbf{Qwen family}}\\
Qwen2.5-VL-72B & \best{0.600} & Qwen2.5-VL-32B & 0.500 \\
Qwen3-VL-235B  & 0.500 & Qwen2.5-VL-7B  & 0.400 \\
Qwen3-VL-30B   & 0.400 & & \\
\midrule
\multicolumn{4}{@{}l}{\textbf{GPT family}}\\
gpt-5-mini & 0.400 & gpt-5-nano & 0.400 \\
\midrule
\multicolumn{4}{@{}l}{\textbf{Gemini family}}\\
gemini-2.5-flash & 0.500 & & \\
\bottomrule
\end{tabular}

\caption{Diagnostic pairwise authenticity discrimination on 10 authentic--imitation pairs in \connoisseurpairs.
A prediction is correct if the model selects the authentic work in each A/B pair.
Given the small scale, results should be interpreted as indicative of a capability gap rather than a definitive model ranking.}
\label{tab:task4_pairs}
\end{table}

Performance remains modest under visually similar confounds: the best models reach 0.60 accuracy (6/10), while most cluster around 0.40--0.50 (RQ4).
Given the diagnostic scale, accuracy estimates carry high variance and we do not draw strong model-ranking conclusions.
Nevertheless, the near-chance performance across all model families suggests that current VLMs struggle with connoisseur-level cues (e.g., brush-and-ink coherence and global style consistency) beyond surface similarity.
We emphasize that \connoisseurpairs{} is not intended to support deployment decisions or commercial authentication; its purpose is to surface this capability gap for future research.
Preliminary expert error attribution (Appendix~\ref{sec:appendix_task4_error}) indicates that errors are consistently associated with weak sensitivity to period/style consistency, overweighting local cues such as inscriptions and seals, and relying on surface-level detail richness rather than global brush-and-ink coherence.

\section{Conclusion}
We introduced \benchmark, a museum-grounded benchmark for evaluating VLMs on Chinese art via four tasks: evidence-grounded QA, structured appreciation, defensible reinterpretation, and diagnostic authenticity discrimination.
Strong models perform well on recognition but degrade on style-to-period inference and hard evidence linking; long-form appreciation and authenticity reasoning remain far from expert performance.
\reinterpret{} and \connoisseurpairs{} are small-scale diagnostic evaluations reflecting expert-curation costs; their conclusions indicate capability gaps rather than definitive rankings.
\benchmark{} highlights these failure modes and provides a unified testbed for improving culturally grounded, evidence-faithful multimodal systems.
Future progress will require stronger visual evidence attribution and more reliable modeling of medium-specific style cues. We hope \benchmark{} supports museum-facing evaluation and motivates broader coverage beyond a single institution.

\section*{Limitations}

\begin{itemize}
  \item \textbf{Limited scale and high curation cost for diagnostic subtasks.}
  \connoisseurpairs{} currently contains only 10 authentic--imitation pairs, and \reinterpret{} covers 25 canonical works.
  Curating \emph{visually close} pairs with defensible provenance and consistent public image availability, or selecting canonical works with established appreciation anchors, requires substantial expert time and careful screening.
  As a result, accuracy estimates have high variance and we treat these tasks as \emph{diagnostic stress tests}; expanding coverage across media (e.g., painting and calligraphy) and imitation types (workshop copies, later replicas, modern reproductions) is necessary for statistically stable conclusions.
  We caution that results from these small-scale tasks should not be used for fine-grained model ranking or as evidence that current systems are suitable for real-world authentication or valuation.

  \item \textbf{LLM-assisted question generation may introduce residual artifacts.}
  \curatorqa{} questions are LLM-assisted and then filtered with constraints and expert review, but artifacts such as recurring templates, skewed distractor patterns, and domain-specific phrasing may remain.
  We mitigate these issues through P1 and P2 separations, visual-answerability checks, a stratified expert audit (Section~3.3), and post hoc validation, but further improvements will benefit from more human-authored questions and adversarial or contrastive construction that explicitly targets shortcut patterns.

  \item \textbf{Single-museum sourcing may limit generalization.}
  \benchmark{} is currently grounded in the Palace Museum catalog and its associated metadata and curatorial writing style.
  While this improves authority and consistency, and while the Palace Museum's collection is exceptionally diverse (spanning over 1.86 million relics across 25 categories and multiple dynasties), performance on \benchmark{} may partly reflect this institution's curatorial framing and catalogue conventions.
  Strong performance may indicate competence on Palace Museum--style data rather than a universally generalizable understanding of Chinese art.
  Future releases should incorporate multi-institution coverage and evaluate cross-museum transfer to disentangle institutional style adaptation from transferable domain reasoning.

  \item \textbf{Automatic metrics only partially capture expert appreciation quality.}
  For \catalogcaption{}, overlap-based metrics and embedding similarity provide scalable signals but do not fully reflect whether an appreciation is truly expert-like, culturally grounded, and well-supported by visual evidence.
  This mismatch is particularly salient when multiple valid interpretations exist or when high-level structure is correct but fine-grained factual units differ.
  We complement automatic metrics with rubric-based expert evaluation (Section~\ref{sec:task2_results}), but future work should further add expert preference judgments and structure-aware evaluation that checks section-wise evidence coverage.
\end{itemize}

\section*{Ethical Considerations}
This section summarizes potential risks and the measures we take to support responsible use of CArtBench.

\noindent\textbf{Potential risks and misuse.} This benchmark evaluates culturally grounded interpretation and diagnostic authenticity discrimination for Chinese artworks. A key risk is overtrust: model outputs may be misused as expert authentication advice, leading to financial or reputational harm in real transactions. A second risk is dual use: insights from the authenticity task could be exploited to improve high-quality imitations. We mitigate these risks by positioning \connoisseurpairs{} explicitly as a diagnostic stress test for benchmarking and failure-mode analysis, requiring cue-based justifications, and emphasizing that outputs should not be used as standalone authentication decisions without expert review. \connoisseurpairs{} is not intended to support deployment decisions, provenance claims, valuation, or certification.

\noindent\textbf{Data sourcing, licensing, and redistribution.}
\benchmark{} is curated from public museum catalog pages and Wikidata entities.
Wikidata content is available under CC0.
Palace Museum images and curatorial texts remain under their respective copyright and terms of use, and we do not redistribute them in the benchmark release.
Instead, we release only stable identifiers and links (e.g., Wikidata QIDs and source URLs) to support attribution and responsible reuse, and users should follow the source sites' policies when accessing or downloading any media.

\noindent\textbf{What we release.}
Our release includes IDs, metadata, prompts, derived annotations, and evaluation code.
We recommend licensing the code under MIT and the derived annotations under CC BY-NC 4.0.
We exclude any third-party copyrighted images or catalog texts from redistribution unless explicit permission is granted by the rights holders.

\noindent\textbf{Privacy and data protection.} The benchmark is derived from museum object pages and is not intended to contain personally identifying information. We screen collected texts for common identifying patterns such as phone numbers, email addresses, and physical addresses, and remove any matches if found. Human ratings are stored in anonymized form and reported only in aggregate.

\noindent\textbf{Cultural context and interpretive validity.} Chinese art interpretation depends on culturally specific concepts and conventions, including medium and format terminology (e.g., \begin{zh}轴\end{zh} `hanging scroll', \begin{zh}卷\end{zh} `handscroll', \begin{zh}册页\end{zh} `album leaf') and style-related norms. Model generations may introduce conceptually misaligned claims or overconfident attributions that are not supported by visible evidence. We therefore emphasize evidence-linked description and require explicit uncertainty when key information cannot be reliably inferred from the image, and our human rubrics penalize unsupported factual assertions and culturally inappropriate interpretations.

\noindent\textbf{Scope and accessibility.} CArtBench is intentionally centered on Chinese art to address the under-representation of East Asian art in current VLM evaluation, which is largely dominated by Western imagery and concepts. We formulate prompts and rubrics in Chinese to reflect the primary language and conventions of the underlying museum sources and real-world curatorial practice, rather than as a claim of culture-agnostic coverage. To support broader use, we provide English translations of prompts and documentation, and we view bilingual and cross-cultural extensions as an important direction for future work.

\section*{Acknowledgements}
X.W.\ acknowledges support from the China Scholarship Council (CSC) program (Project No.\ 202308070072), X.Z.\ acknowledges support from the CSC program (Project No.\ 202308070070), and Z.W.\ acknowledges support from the CSC program (Project No.\ 202408210062) and the Dalian Federation of Social Sciences Associations (Project No.\ 2026dlskzd051).

\bibliography{custom}

@article{vrandecic2014wikidata,
  title   = {Wikidata: a free collaborative knowledgebase},
  author  = {Vrande{\v{c}}i{\'c}, Denny and Kr{\"o}tzsch, Markus},
  journal = {Communications of the ACM},
  volume  = {57},
  number  = {10},
  pages   = {78--85},
  year    = {2014},
  doi     = {10.1145/2629489},
  url     = {https://doi.org/10.1145/2629489}
}

@inproceedings{jin-etal-2023-museumqa,
  title     = {MuseumQA: A Fine-grained Question Answering Dataset for Museums and Artifacts},
  author    = {Jin, Feng and Chang, Qingling and Xu, Zehua},
  booktitle = {Proceedings of the 2023 6th International Conference on Machine Learning and Natural Language Processing (MLNLP '23)},
  year      = {2023},
  pages     = {221--226},
  publisher = {Association for Computing Machinery},
  address   = {New York, NY, USA},
  doi       = {10.1145/3639479.3639525},
  url       = {https://doi.org/10.1145/3639479.3639525}
}

@article{geirhos2020shortcut,
  title   = {Shortcut Learning in Deep Neural Networks},
  author  = {Geirhos, Robert and Jacobsen, J{\"o}rn-Henrik and Michaelis, Claudio and Zemel, Richard and Brendel, Wieland and Bethge, Matthias and Wichmann, Felix A.},
  journal = {Nature Machine Intelligence},
  volume  = {2},
  number  = {11},
  pages   = {665--673},
  year    = {2020},
  doi     = {10.1038/s42256-020-00257-z},
  url     = {https://www.nature.com/articles/s42256-020-00257-z}
}

@inproceedings{agrawal2018dontjustassume,
  title     = {Don't Just Assume; Look and Answer: Overcoming Priors for Visual Question Answering},
  author    = {Agrawal, Aishwarya and Batra, Dhruv and Parikh, Devi and Kembhavi, Aniruddha},
  booktitle = {Proceedings of the IEEE/CVF Conference on Computer Vision and Pattern Recognition (CVPR)},
  year      = {2018},
  pages     = {4971--4980},
  doi       = {10.1109/CVPR.2018.00522},
  url       = {https://openaccess.thecvf.com/content_cvpr_2018/html/Agrawal_Dont_Just_Assume_CVPR_2018_paper.html}
}

@misc{ananthram2024seeit,
  title         = {See It from My Perspective: How Language Affects Cultural Bias in Image Understanding},
  author        = {Ananthram, Amith and Stengel-Eskin, Elias and Vondrick, Carl and Bansal, Mohit and McKeown, Kathleen},
  year          = {2024},
  eprint        = {2406.11665},
  archivePrefix = {arXiv},
  primaryClass  = {cs.CV},
  url           = {https://arxiv.org/abs/2406.11665}
}

@inproceedings{li-etal-2023-evaluating,
  title     = {Evaluating Object Hallucination in Large Vision-Language Models},
  author    = {Li, Yifan and Du, Yifan and Zhou, Kun and Wang, Jinpeng and Zhao, Xin and Wen, Ji-Rong},
  booktitle = {Proceedings of the 2023 Conference on Empirical Methods in Natural Language Processing},
  month     = {dec},
  year      = {2023},
  address   = {Singapore},
  publisher = {Association for Computational Linguistics},
  url       = {https://aclanthology.org/2023.emnlp-main.20/},
  doi       = {10.18653/v1/2023.emnlp-main.20},
  pages     = {292--305}
}

@article{chen2025acpas,
  title   = {ACPAS: an expert-assistance system for authenticating ancient Chinese paintings via LLM-based agents},
  author  = {Chen, Xiaojiao and Li, Yueying and Chen, Yonghao and Tang, Tan and Wang, Ruihan and Wang, Yifan and Feng, Yingchaojie and Chen, Wei and Wang, Xiaosong},
  journal = {npj Heritage Science},
  volume  = {13},
  pages   = {512},
  year    = {2025},
  doi     = {10.1038/s40494-025-02093-z},
  url     = {https://www.nature.com/articles/s40494-025-02093-z}
}

@inproceedings{antol2015vqa,
  title     = {VQA: Visual Question Answering},
  author    = {Antol, Stanislaw and Agrawal, Aishwarya and Lu, Jiasen and Mitchell, Margaret and Batra, Dhruv and Lawrence Zitnick, C. and Parikh, Devi},
  booktitle = {Proceedings of the IEEE International Conference on Computer Vision (ICCV)},
  year      = {2015},
  pages     = {2425--2433},
  doi       = {10.1109/ICCV.2015.279},
  url       = {https://openaccess.thecvf.com/content_iccv_2015/html/Antol_VQA_Visual_Question_ICCV_2015_paper.html}
}

@inproceedings{lin2014mscoco,
  title     = {Microsoft COCO: Common Objects in Context},
  author    = {Lin, Tsung-Yi and Maire, Michael and Belongie, Serge and Hays, James and Perona, Pietro and Ramanan, Deva and Doll{\'a}r, Piotr and Zitnick, C. Lawrence},
  booktitle = {European Conference on Computer Vision (ECCV)},
  year      = {2014},
  pages     = {740--755},
  publisher = {Springer},
  doi       = {10.1007/978-3-319-10602-1_48}
}

@misc{cvlue2024,
  title         = {CVLUE: A New Benchmark Dataset for Chinese Vision-Language Understanding Evaluation},
  author        = {Wang, Yuxuan and Liu, Yijun and Yu, Fei and Huang, Chen and Li, Kexin and Wan, Zhiguo and Che, Wanxiang},
  year          = {2024},
  eprint        = {2407.01081},
  archivePrefix = {arXiv},
  primaryClass  = {cs.CV},
  doi           = {10.48550/arXiv.2407.01081},
  url           = {https://arxiv.org/abs/2407.01081}
}

@inproceedings{nayak-etal-2024-benchmarking,
  title     = {Benchmarking Vision Language Models for Cultural Understanding},
  author    = {Nayak, Shravan and Jain, Kanishk and Awal, Rabiul and Reddy, Siva and Steenkiste, Sjoerd Van and Hendricks, Lisa Anne and Stanczak, Karolina and Agrawal, Aishwarya},
  booktitle = {Proceedings of the 2024 Conference on Empirical Methods in Natural Language Processing},
  month     = nov,
  year      = {2024},
  address   = {Miami, Florida, USA},
  publisher = {Association for Computational Linguistics},
  pages     = {5769--5790},
  doi       = {10.18653/v1/2024.emnlp-main.329},
  url       = {https://aclanthology.org/2024.emnlp-main.329/}
}

@InProceedings{Achlioptas_2021_CVPR,
  author    = {Achlioptas, Panos and Ovsjanikov, Maks and Haydarov, Kilichbek and Elhoseiny, Mohamed and Guibas, Leonidas J.},
  title     = {ArtEmis: Affective Language for Visual Art},
  booktitle = {Proceedings of the IEEE/CVF Conference on Computer Vision and Pattern Recognition (CVPR)},
  month     = {June},
  year      = {2021},
  pages     = {11569--11579},
  url       = {https://openaccess.thecvf.com/content/CVPR2021/html/Achlioptas_ArtEmis_Affective_Language_for_Visual_Art_CVPR_2021_paper.html}
}

@article{wan2025chpcqa,
  title   = {Constructing a large-scale {QA} system for Chinese painting and calligraphy},
  author  = {Wan, Jing and Li, Xinrong and Zhang, Hao and Zou, Ao and Wang, Rumei},
  journal = {npj Heritage Science},
  year    = {2025},
  doi     = {10.1038/s40494-025-02219-3},
  url     = {https://www.nature.com/articles/s40494-025-02219-3}
}

@article{chao2024arteyer,
  title   = {ArtEyer: A multimodal analysis and authentication system with contextual data visualizations for fine art},
  author  = {Chao, Wen and Chen, Zhihua and Yuan, Linping},
  journal = {Visual Informatics},
  volume  = {14},
  number  = {4},
  pages   = {134--144},
  year    = {2024},
  doi     = {10.1016/j.visinf.2024.11.001},
  url     = {https://www.sciencedirect.com/science/article/pii/S2468502X24001023}
}

@misc{openai_gpt52_model,
  author       = {OpenAI},
  title        = {GPT-5.2 Model},
  year         = {2025},
  howpublished = {\url{https://platform.openai.com/docs/models/gpt-5.2}},
  note         = {Accessed: 2026-01-05}
}

@article{bertscore,
  title={BERTScore: Evaluating Text Generation with BERT},
  author={Zhang, Tianyi and Kishore, Varsha and Wu, Felix and Weinberger, Kilian Q. and Artzi, Yoav},
  journal={arXiv preprint arXiv:1904.09675},
  year={2019}
}

@inproceedings{rouge,
  title={ROUGE: A Package for Automatic Evaluation of Summaries},
  author={Lin, Chin-Yew},
  booktitle={Text Summarization Branches Out},
  pages={74--81},
  year={2004},
  publisher={Association for Computational Linguistics}
}

@inproceedings{bleu,
  title={Bleu: a Method for Automatic Evaluation of Machine Translation},
  author={Papineni, Kishore and Roukos, Salim and Ward, Todd and Zhu, Wei-Jing},
  booktitle={Proceedings of the 40th Annual Meeting of the Association for Computational Linguistics},
  pages={311--318},
  year={2002}
}

@inproceedings{cider,
  title={CIDEr: Consensus-based Image Description Evaluation},
  author={Vedantam, Ramakrishna and Zitnick, C. Lawrence and Parikh, Devi},
  booktitle={Proceedings of CVPR},
  year={2015}
}

@book{torrance1974ttct,
  title     = {Torrance Tests of Creative Thinking: Norms-Technical Manual},
  author    = {Torrance, E. Paul},
  year      = {1974},
  publisher = {Ginn and Company},
  address   = {Lexington, MA}
}

@misc{qw25vl,
      title={Qwen2.5-VL Technical Report}, 
      author={Shuai Bai and Keqin Chen and Xuejing Liu and Jialin Wang and Wenbin Ge and Sibo Song and Kai Dang and Peng Wang and Shijie Wang and Jun Tang and Humen Zhong and Yuanzhi Zhu and Mingkun Yang and Zhaohai Li and Jianqiang Wan and Pengfei Wang and Wei Ding and Zheren Fu and Yiheng Xu and Jiabo Ye and Xi Zhang and Tianbao Xie and Zesen Cheng and Hang Zhang and Zhibo Yang and Haiyang Xu and Junyang Lin},
      year={2025},
      eprint={2502.13923},
      archivePrefix={arXiv},
      primaryClass={cs.CV},
      url={https://arxiv.org/abs/2502.13923}, 
}

@misc{qw3vl,
      title={Qwen3-VL Technical Report}, 
      author={Shuai Bai and Yuxuan Cai and Ruizhe Chen and Keqin Chen and Xionghui Chen and Zesen Cheng and Lianghao Deng and Wei Ding and Chang Gao and Chunjiang Ge and Wenbin Ge and Zhifang Guo and Qidong Huang and Jie Huang and Fei Huang and Binyuan Hui and Shutong Jiang and Zhaohai Li and Mingsheng Li and Mei Li and Kaixin Li and Zicheng Lin and Junyang Lin and Xuejing Liu and Jiawei Liu and Chenglong Liu and Yang Liu and Dayiheng Liu and Shixuan Liu and Dunjie Lu and Ruilin Luo and Chenxu Lv and Rui Men and Lingchen Meng and Xuancheng Ren and Xingzhang Ren and Sibo Song and Yuchong Sun and Jun Tang and Jianhong Tu and Jianqiang Wan and Peng Wang and Pengfei Wang and Qiuyue Wang and Yuxuan Wang and Tianbao Xie and Yiheng Xu and Haiyang Xu and Jin Xu and Zhibo Yang and Mingkun Yang and Jianxin Yang and An Yang and Bowen Yu and Fei Zhang and Hang Zhang and Xi Zhang and Bo Zheng and Humen Zhong and Jingren Zhou and Fan Zhou and Jing Zhou and Yuanzhi Zhu and Ke Zhu},
      year={2025},
      eprint={2511.21631},
      archivePrefix={arXiv},
      primaryClass={cs.CV},
      url={https://arxiv.org/abs/2511.21631}, 
}

@misc{glm45v,
      title={GLM-4.5V and GLM-4.1V-Thinking: Towards Versatile Multimodal Reasoning with Scalable Reinforcement Learning}, 
      author={V Team and Wenyi Hong and Wenmeng Yu and Xiaotao Gu and Guo Wang and Guobing Gan and Haomiao Tang and Jiale Cheng and Ji Qi and Junhui Ji and Lihang Pan and Shuaiqi Duan and Weihan Wang and Yan Wang and Yean Cheng and Zehai He and Zhe Su and Zhen Yang and Ziyang Pan and Aohan Zeng and Baoxu Wang and Bin Chen and Boyan Shi and Changyu Pang and Chenhui Zhang and Da Yin and Fan Yang and Guoqing Chen and Haochen Li and Jiale Zhu and Jiali Chen and Jiaxing Xu and Jiazheng Xu and Jing Chen and Jinghao Lin and Jinhao Chen and Jinjiang Wang and Junjie Chen and Leqi Lei and Letian Gong and Leyi Pan and Mingdao Liu and Mingde Xu and Mingzhi Zhang and Qinkai Zheng and Ruiliang Lyu and Shangqin Tu and Sheng Yang and Shengbiao Meng and Shi Zhong and Shiyu Huang and Shuyuan Zhao and Siyan Xue and Tianshu Zhang and Tianwei Luo and Tianxiang Hao and Tianyu Tong and Wei Jia and Wenkai Li and Xiao Liu and Xiaohan Zhang and Xin Lyu and Xinyu Zhang and Xinyue Fan and Xuancheng Huang and Yadong Xue and Yanfeng Wang and Yanling Wang and Yanzi Wang and Yifan An and Yifan Du and Yiheng Huang and Yilin Niu and Yiming Shi and Yu Wang and Yuan Wang and Yuanchang Yue and Yuchen Li and Yusen Liu and Yutao Zhang and Yuting Wang and Yuxuan Zhang and Zhao Xue and Zhengxiao Du and Zhenyu Hou and Zihan Wang and Peng Zhang and Debing Liu and Bin Xu and Juanzi Li and Minlie Huang and Yuxiao Dong and Jie Tang},
      year={2026},
      eprint={2507.01006},
      archivePrefix={arXiv},
      primaryClass={cs.CV},
      url={https://arxiv.org/abs/2507.01006}, 
}

@misc{openai_gpt5,
  title        = {GPT-5 Model},
  author       = {{OpenAI}},
  howpublished = {OpenAI API Documentation},
  year         = {2026},
  note         = {Accessed 2026-01-05},
  url          = {https://platform.openai.com/docs/models/gpt-5}
}

@misc{openai_gpt52,
  title        = {GPT-5.2 Model},
  author       = {{OpenAI}},
  howpublished = {OpenAI API Documentation},
  year         = {2026},
  note         = {Accessed 2026-01-05},
  url          = {https://platform.openai.com/docs/models/gpt-5.2}
}

@misc{openai_gpt5mini,
  title        = {GPT-5 mini Model},
  author       = {{OpenAI}},
  howpublished = {OpenAI API Documentation},
  year         = {2026},
  note         = {Accessed 2026-01-05},
  url          = {https://platform.openai.com/docs/models/gpt-5-mini}
}

@misc{openai_gpt5nano,
  title        = {GPT-5 nano Model},
  author       = {{OpenAI}},
  howpublished = {OpenAI API Documentation},
  year         = {2026},
  note         = {Accessed 2026-01-05},
  url          = {https://platform.openai.com/docs/models/gpt-5-nano}
}

@misc{google_gemini25flash,
  title        = {Gemini 2.5 Flash Model},
  author       = {{Google}},
  howpublished = {Google AI for Developers: Gemini API Documentation},
  year         = {2025},
  note         = {Accessed 2026-01-05},
  url          = {https://ai.google.dev/gemini-api/docs/models/gemini}
}

@misc{guan2024hallusionbenchadvanceddiagnosticsuite,
      title={HallusionBench: An Advanced Diagnostic Suite for Entangled Language Hallucination and Visual Illusion in Large Vision-Language Models}, 
      author={Tianrui Guan and Fuxiao Liu and Xiyang Wu and Ruiqi Xian and Zongxia Li and Xiaoyu Liu and Xijun Wang and Lichang Chen and Furong Huang and Yaser Yacoob and Dinesh Manocha and Tianyi Zhou},
      year={2024},
      eprint={2310.14566},
      archivePrefix={arXiv},
      primaryClass={cs.CV},
      url={https://arxiv.org/abs/2310.14566}, 
}

@inproceedings{fu2025mme,
  title={Mme: A comprehensive evaluation benchmark for multimodal large language models},
  author={Fu, Chaoyou and Chen, Peixian and Shen, Yunhang and Qin, Yulei and Zhang, Mengdan and Lin, Xu and Yang, Jinrui and Zheng, Xiawu and Li, Ke and Sun, Xing and others},
  booktitle={The Thirty-ninth Annual Conference on Neural Information Processing Systems Datasets and Benchmarks Track},
  year={2025}
}

@article{yu2023mm,
  title={Mm-vet: Evaluating large multimodal models for integrated capabilities},
  author={Yu, Weihao and Yang, Zhengyuan and Li, Linjie and Wang, Jianfeng and Lin, Kevin and Liu, Zicheng and Wang, Xinchao and Wang, Lijuan},
  journal={arXiv preprint arXiv:2308.02490},
  year={2023}
}

@inproceedings{ozaki2025towards,
  title={Towards cross-lingual explanation of artwork in large-scale vision language models},
  author={Ozaki, Shintaro and Hayashi, Kazuki and Sakai, Yusuke and Kamigaito, Hidetaka and Hayashi, Katsuhiko and Watanabe, Taro},
  booktitle={Findings of the Association for Computational Linguistics: NAACL 2025},
  pages={3773--3809},
  year={2025}
}

@inproceedings{mohamed2024no,
  title={No culture left behind: ArtELingo-28, a benchmark of WikiArt with captions in 28 languages},
  author={Mohamed, Youssef and Li, Runjia and Ahmad, Ibrahim Sa’id and Haydarov, Kilichbek and Torr, Philip and Church, Kenneth and Elhoseiny, Mohamed},
  booktitle={Proceedings of the 2024 Conference on Empirical Methods in Natural Language Processing},
  pages={20939--20962},
  year={2024}
}

@inproceedings{lan2025f2bench,
  title={F$^2$Bench: An Open-ended Fairness Evaluation Benchmark for LLMs with Factuality Considerations},
  author={Lan, Tian and Li, Jiang and Wang, Yemin and Liu, Xu and Su, Xiangdong and Gao, Guanglai},
  booktitle={Proceedings of the 2025 Conference on Empirical Methods in Natural Language Processing},
  pages={2031--2046},
  year={2025}
}

@inproceedings{lan2025mcbe,
  title={McBE: A Multi-task Chinese Bias Evaluation Benchmark for Large Language Models},
  author={Lan, Tian and Su, Xiangdong and Liu, Xu and Wang, Ruirui and Chang, Ke and Li, Jiang and Gao, Guanglai},
  booktitle={Findings of the Association for Computational Linguistics: ACL 2025},
  pages={6033--6056},
  year={2025}
}

@misc{he2024cmmu,
      title={CMMU: A Benchmark for Chinese Multi-modal Multi-type Question Understanding and Reasoning}, 
      author={Zheqi He and Xinya Wu and Pengfei Zhou and Richeng Xuan and Guang Liu and Xi Yang and Qiannan Zhu and Hua Huang},
      year={2024},
      eprint={2401.14011},
      archivePrefix={arXiv},
      primaryClass={cs.CL},
      url={https://arxiv.org/abs/2401.14011}, 
}

@misc{zhang2024cmmmu,
      title={CMMMU: A Chinese Massive Multi-discipline Multimodal Understanding Benchmark}, 
      author={Ge Zhang and Xinrun Du and Bei Chen and Yiming Liang and Tongxu Luo and Tianyu Zheng and Kang Zhu and Yuyang Cheng and Chunpu Xu and Shuyue Guo and Haoran Zhang and Xingwei Qu and Junjie Wang and Ruibin Yuan and Yizhi Li and Zekun Wang and Yudong Liu and Yu-Hsuan Tsai and Fengji Zhang and Chenghua Lin and Wenhao Huang and Jie Fu},
      year={2024},
      eprint={2401.11944},
      archivePrefix={arXiv},
      primaryClass={cs.CL},
      url={https://arxiv.org/abs/2401.11944}, 
}

@article{xu2024muraldh,
  author = {Xu, Zishan and Yang, Yuqing and Fang, Qianzhen and Chen, Wei and Xu, Tingting and Liu, Jueting and Wang, Zehua},
  title = {A comprehensive dataset for digital restoration of Dunhuang murals},
  journal = {Scientific Data},
  volume = {11},
  pages = {955},
  year = {2024},
  doi = {10.1038/s41597-024-03785-0},
  url = {https://doi.org/10.1038/s41597-024-03785-0}
}

@article{liu2025deepjiandu,
  author = {Liu, Yiran and Zhang, Qiang and Qi, Ying and Wan, Teng and Zhang, Defang and Li, Yutong and Zhang, Xin and Ma, Longbin and Ruan, Qiuyue and Guo, Huanting and Li, Yingchun and Miao, Xinyue and Xiao, Wenjun and Li, Yongbo and Yuan, Jiang and Chen, Shanxiong},
  title = {DeepJiandu Dataset for Character Detection and Recognition on Jiandu Manuscript},
  journal = {Scientific Data},
  volume = {12},
  pages = {398},
  year = {2025},
  doi = {10.1038/s41597-025-04716-3},
  url = {https://doi.org/10.1038/s41597-025-04716-3}
}

@misc{zhao2025mccd,
      title={MCCD: A Multi-Attribute Chinese Calligraphy Character Dataset Annotated with Script Styles, Dynasties, and Calligraphers}, 
      author={Yixin Zhao and Yuyi Zhang and Lianwen Jin},
      year={2025},
      eprint={2507.06948},
      archivePrefix={arXiv},
      primaryClass={cs.CV},
      url={https://arxiv.org/abs/2507.06948}, 
}

\appendix

\section{Appendix}
\label{sec:appendix}

\section{Additional Analysis}
\label{sec:appendix_extra}

\begin{table*}[t]
\centering
\small
\setlength{\tabcolsep}{10pt}
\renewcommand{\arraystretch}{1.12}
\begin{tabularx}{\textwidth}{@{}l l c c c >{\raggedright\arraybackslash}X@{}}
\toprule
\textbf{Bilingual model} & \textbf{General model} & $\mathbf{Acc.}_{\text{bi}}$ & $\mathbf{Acc.}_{\text{gen}}$ & $\boldsymbol{\Delta}$ & \textbf{Significance} \\
\midrule
Qwen3-235B & GPT-5-mini & 0.8400 & 0.7949 & +0.0451 &
McNemar (paired, $N{=}14421$): $p{=}4.22{\times}10^{-47}$.\newline
Clustered by artwork ID ($N_{\text{artwork}}{=}1589$): $\overline{\Delta}{=}0.0447$,
CI$_{95}{=}[0.0384,\,0.0512]$, $p{=}0.0002$. \\
\bottomrule
\end{tabularx}
\caption{Chinese--English bilingual vs.\ general-purpose API proxy comparison on \curatorqa. Qwen3-235B is a Chinese--English bilingual VLM with strong Chinese multimodal performance; GPT-5-mini is a general-purpose multilingual API.}
\label{tab:curatorqa-lang-significance}
\end{table*}
\subsection{KPI Weight Sensitivity Analysis for \catalogcaption}
\label{sec:appendix_task2_kpi_sensitivity}

\paragraph{Motivation.}
KPI aggregates multiple automatic metrics (BERTScore$_{F1}$, CIDEr-like, ROUGE-L, BLEU-4) into a single score
(Equation~\ref{eq:kpi}). Since any fixed weighting may appear arbitrary, we test whether our \catalogcaption{}
model ranking is stable under reasonable perturbations of the metric weights.

\paragraph{Protocol.}
Let $\mathbf{w}=(w_{\text{bert}}, w_{\text{cider}}, w_{\text{rouge}}, w_{\text{bleu}})$ be a nonnegative weight vector
with $\sum_i w_i = 1$. For each sampled $\mathbf{w}$, we compute
\(
\mathrm{KPI}_{\mathbf{w}} = \sum_i w_i\,m_i
\)
where $m_i$ denotes the corresponding metric value, then obtain a model ranking induced by $\mathrm{KPI}_{\mathbf{w}}$.
We compare each ranking to the default-weight ranking using Spearman's $\rho$ and Kendall's $\tau$.
We also compute (i) \emph{Top-1 same rate}, the fraction of scenarios in which the best-performing \emph{model}
(excluding human baselines) is unchanged, and (ii) \emph{Top-3 Jaccard}, the Jaccard similarity between the
default top-3 model set and the top-3 set under $\mathbf{w}$.

\paragraph{Weight perturbation schemes.}
We consider the following perturbations:
(i) \textbf{Uniform}: $w_i = 0.25$ for all metrics.
(ii) \textbf{Dirichlet}: $\mathbf{w} \sim \mathrm{Dirichlet}(\alpha\mathbf{1})$ with $\alpha\in\{0.5,1,5\}$,
capturing increasingly concentrated vs.\ smooth weight distributions.
(iii) \textbf{Local Gaussian}: sample $\tilde{\mathbf{w}} \sim \mathcal{N}(\mathbf{w}_0, \sigma^2 I)$ around the
default weight $\mathbf{w}_0=(0.45,0.25,0.20,0.10)$ with $\sigma=0.05$, then project onto the simplex.
(iv) \textbf{BERT-dominant range}: sample $w_{\text{bert}}\in[0.35,0.60]$ and distribute the remaining mass across
other metrics proportional to the default weights.

\begin{table*}[t]
\centering
\small
\setlength{\tabcolsep}{1pt}
\renewcommand{\arraystretch}{1.08}
\begin{tabular}{l r c c c c}
\toprule
\textbf{Scenario} & \textbf{\#samples} &
\textbf{Spearman (p05/p50/p95)} &
\textbf{Kendall (p05/p50/p95)} &
\textbf{Top-1 same} &
\textbf{Top-3 Jaccard (p05/p50/p95)} \\
\midrule
Default (sanity) & 1 &
1.000/1.000/1.000 &
1.000/1.000/1.000 &
1.000 &
1.000/1.000/1.000 \\
Uniform weights & 1 &
0.991/0.991/0.991 &
0.964/0.964/0.964 &
1.000 &
1.000/1.000/1.000 \\
Dirichlet $\alpha{=}5$ & 8000 &
0.945/0.991/0.991 &
0.855/0.964/0.964 &
1.000 &
1.000/1.000/1.000 \\
Dirichlet $\alpha{=}1$ & 8000 &
0.882/0.964/1.000 &
0.709/0.891/1.000 &
1.000 &
0.500/1.000/1.000 \\
Dirichlet $\alpha{=}0.5$ & 8000 &
0.873/0.964/1.000 &
0.673/0.891/1.000 &
1.000 &
0.500/1.000/1.000 \\
$w_{\text{bert}}\!\in\![0.35,0.60]$ & 5000 &
0.991/1.000/1.000 &
0.964/1.000/1.000 &
1.000 &
1.000/1.000/1.000 \\
Local Gaussian $\sigma{=}0.05$ & 5000 &
0.982/0.991/1.000 &
0.927/0.964/1.000 &
1.000 &
0.500/1.000/1.000 \\
\bottomrule
\end{tabular}
\caption{KPI weight sensitivity analysis for \catalogcaption.
Rank correlations are computed against the default-weight ordering; we report 5th/50th/95th percentiles for stochastic scenarios.
Top-1 same and Top-3 Jaccard are computed over \emph{models} (excluding human baselines).}
\label{tab:appendix_task2_kpi_sensitivity}
\end{table*}

\paragraph{Findings.}
As shown in Table~\ref{tab:appendix_task2_kpi_sensitivity}, rankings remain highly consistent with the default
ordering across a wide range of weight perturbations (median Spearman $\ge 0.964$; median Kendall $\ge 0.891$ even
under the more extreme Dirichlet sampling with $\alpha\le 1$).
Crucially, the top-ranked model under the default KPI remains unchanged across all tested scenarios
(Top-1 same $=1.0$), supporting that our main \catalogcaption conclusions do not hinge on a particular weighting choice.

\begin{table}[t]
\centering
\small
\setlength{\tabcolsep}{3pt}
\renewcommand{\arraystretch}{1.06}
\begin{tabular}{l c c c}
\toprule
\textbf{Model} & \textbf{Mean rank} & \textbf{Std. rank} & \textbf{$P(\text{top-3})$} \\
\midrule
human\_mean & 1.000 & 0.000 & 1.000 \\
preds\_qwen25\_72B & 2.397 & 0.734 & 0.860 \\
preds\_qwen3\_235B & 2.863 & 0.495 & 0.936 \\
preds\_qwen3\_30B & 4.035 & 0.806 & 0.203 \\
preds\_qwen25\_7B & 5.038 & 0.920 & 0.000 \\
preds\_glm45 & 6.189 & 0.785 & 0.000 \\
preds\_glm46 & 6.764 & 0.766 & 0.000 \\
preds\_qwen25\_32B & 8.373 & 1.083 & 0.000 \\
gpt5nano & 8.733 & 0.860 & 0.000 \\
gpt5mini & 9.608 & 0.509 & 0.000 \\
gemini2.5 & 11.000 & 0.000 & 0.000 \\
\bottomrule
\end{tabular}
\caption{Model rank stability under Dirichlet weight perturbations with $\alpha{=}1$ (8{,}000 samples).
We report the mean and standard deviation of each model's rank, and the probability of appearing in the top-3.}
\label{tab:appendix_task2_rank_stability}
\end{table}

\subsection{Complementary Expert Evaluation for \catalogcaption}
\label{sec:appendix_task2_expert}

To validate that automatic metrics are not solely driven by institutional style overlap, we conducted a rubric-based expert evaluation on 258 model--artwork pairs (86 artworks $\times$ 3 representative models: Qwen2.5-VL-72B, Qwen3-VL-235B, and Gemini-2.5-flash).
Three raters scored each output on D1 (Evidence-grounded Interpretation), D2 (Cultural/Art-historical Appropriateness), and D3 (Insight Depth) using a 1--5 Likert scale, and flagged major factual/hallucination errors with a binary indicator (E1).
Table~\ref{tab:appendix_task2_expert_eval} reports the results.
The 3-rater mean shows moderate reliability (ICC(2,3) $\approx$ 0.62--0.64 for D1--D3).
For E1, Cohen's $\kappa$ is low (0.012) due to the rarity of major errors, but Gwet's AC1 (0.841) and raw agreement (0.863) confirm high consistency on this near-ceiling item.

\begin{table}[t]
\centering
\small
\setlength{\tabcolsep}{3pt}
\renewcommand{\arraystretch}{1.06}
\begin{tabular}{l c c c c}
\toprule
\textbf{Item} & \textbf{D1} & \textbf{D2} & \textbf{D3} & \textbf{E1} \\
\midrule
\textbf{Agreement} & .639 & .622 & .637 & .841$^\dagger$ \\
\midrule
Qwen2.5-72B & 3.640 & 3.702 & 3.663 & 0.081 \\
Qwen3-235B  & 3.647 & 3.609 & 3.616 & 0.050 \\
Gemini-flash & 3.229 & 3.275 & 3.337 & 0.093 \\
\bottomrule
\end{tabular}
\caption{Expert rubric evaluation for \catalogcaption. Agreement: ICC(2,3) for D1--D3; $^\dagger$Gwet's AC1 for E1. E1 = fraction of outputs flagged for major factual errors.}
\label{tab:appendix_task2_expert_eval}
\end{table}

\subsection{Inter-Annotator Agreement for \reinterpret}
\label{sec:appendix_task3_iaa}

Table~\ref{tab:appendix_task3_iaa_table} reports inter-annotator agreement statistics for \reinterpret.
Stage~1 (plausibility gate) agreement is measured by Fleiss' $\kappa$; Stage~2 agreement is measured by ICC(2,3) on the 3-rater mean.

\begin{table}[t]
\centering
\small
\setlength{\tabcolsep}{2pt}
\renewcommand{\arraystretch}{1.06}
\begin{tabular}{l c c c}
\toprule
\textbf{Model} & \textbf{Gate $\kappa$} & \textbf{ICC Primary} & \textbf{ICC All} \\
\midrule
GLM-4.6V         & 0.487 & 0.498 & 0.649 \\
gemini-2.5-flash & 0.415 & 0.476 & 0.709 \\
gpt-5-mini       & 0.711 & 0.785 & 0.869 \\
qwen3-vl-235B    & 0.577 & 0.877 & 0.888 \\
\bottomrule
\end{tabular}
\caption{Inter-annotator agreement for \reinterpret. Gate $\kappa$: Fleiss' $\kappa$ for Stage~1 pass/fail. ICC Primary/All: ICC(2,3) for the 3-rater mean on D1--D3 and D1--D5 respectively.}
\label{tab:appendix_task3_iaa_table}
\end{table}

\subsection{Additional statistics for \reinterpret}
\label{sec:appendix_task3_stats}

We provide additional breakdowns for \reinterpret, including dimension-wise macro-averages
with bootstrapped 95\% confidence intervals and Stage~1 gate-failure type distributions.

\begin{table*}[t]
\centering
\scriptsize
\setlength{\tabcolsep}{2pt}
\renewcommand{\arraystretch}{1.15}
\begin{tabular}{p{0.17\textwidth} p{0.11\textwidth} p{0.14\textwidth} p{0.14\textwidth} p{0.14\textwidth} p{0.14\textwidth} p{0.14\textwidth}}
\toprule
\textbf{Model} &
\textbf{Gate pass} &
\textbf{D1 Novelty} &
\textbf{D2 Coherence} &
\textbf{D3 Evidence} &
\textbf{D4 Elaboration} &
\textbf{D5 Insight} \\
\midrule
gemini-2.5-flash &
0.707 [0.600, 0.813] &
3.287 [2.940, 3.653] &
3.960 [3.740, 4.193] &
3.773 [3.587, 3.960] &
3.980 [3.760, 4.213] &
3.613 [3.353, 3.907] \\
GLM-4.6V &
0.853 [0.760, 0.933] &
3.407 [3.213, 3.620] &
3.793 [3.653, 3.947] &
3.747 [3.560, 3.947] &
3.800 [3.627, 3.980] &
3.647 [3.467, 3.853] \\
gpt-5-mini &
0.867 [0.787, 0.947] &
3.280 [3.120, 3.447] &
3.767 [3.647, 3.880] &
3.653 [3.507, 3.787] &
3.873 [3.733, 4.027] &
3.513 [3.333, 3.700] \\
qwen3-vl-235B &
0.880 [0.787, 0.960] &
3.533 [3.340, 3.740] &
3.867 [3.687, 4.053] &
3.840 [3.707, 3.973] &
3.847 [3.693, 4.013] &
3.553 [3.413, 3.700] \\
\bottomrule
\end{tabular}
\caption{Dimension-wise macro-averages for \reinterpret\ with 95\% bootstrap confidence intervals over artworks.
Gate pass is the Stage~1 plausibility pass rate.
D1--D5 are Stage~2 Likert ratings computed only for gate-passed outputs, macro-averaged over artworks.}
\label{tab:appendix_task3_dimwise}
\end{table*}

\begin{table*}[t]
\centering
\scriptsize
\setlength{\tabcolsep}{2pt}
\renewcommand{\arraystretch}{1.15}
\begin{tabular}{p{0.19\textwidth} p{0.10\textwidth} p{0.14\textwidth} p{0.14\textwidth} p{0.14\textwidth} p{0.16\textwidth} p{0.10\textwidth}}
\toprule
\textbf{Model} &
\textbf{Failures} &
\textbf{Other} &
\textbf{Era/Dynasty error} &
\textbf{Misread / missed content} &
\textbf{Wrong type/region attribution} &
\textbf{Empty/NA} \\
\midrule
gemini-2.5-flash &
22 (29.3\%) &
11 (14.7\%) &
6 (8.0\%) &
2 (2.7\%) &
2 (2.7\%) &
1 (1.3\%) \\
GLM-4.6V &
11 (14.7\%) &
8 (10.7\%) &
2 (2.7\%) &
1 (1.3\%) &
0 (0.0\%) &
0 (0.0\%) \\
gpt-5-mini &
10 (13.3\%) &
5 (6.7\%) &
1 (1.3\%) &
3 (4.0\%) &
1 (1.3\%) &
0 (0.0\%) \\
qwen3-vl-235B &
9 (12.0\%) &
7 (9.3\%) &
1 (1.3\%) &
0 (0.0\%) &
1 (1.3\%) &
0 (0.0\%) \\
\bottomrule
\end{tabular}
\caption{Stage~1 gate-failure type distributions for \reinterpret.
Counts are over rater$\times$artwork instances (75 per model), with percentages relative to 75.
Failure types are recorded only when a rater marks the output as failing the plausibility gate.}
\label{tab:appendix_task3_failtypes}
\end{table*}

\subsection{Human Evaluation Protocol and Annotator Information}
Human evaluations were conducted by three domain experts (two art graduate students and one art-school professor in China), all adult native Chinese speakers with formal training in Chinese art history. Annotators were compensated at a reasonable rate meeting or exceeding local minimum wage or typical market rates. Participation was voluntary with research-use notice; we collected no personally identifying information and report only anonymized, aggregate results.

\subsection{Preliminary expert error attribution for \connoisseurpairs}
\label{sec:appendix_task4_error}
We summarize a preliminary expert attribution on misclassified cases in \connoisseurpairs.
Errors are consistently associated with weak sensitivity to period/style consistency, overweighting local cues
such as inscriptions and seals, and relying on surface-level detail richness rather than global brush-and-ink coherence.
Given the diagnostic scale (10 pairs), these observations should be treated as indicative patterns rather than statistically validated conclusions.

\subsection{Reproducibility Details}

\subsection{Prompt for \curatorqa}
\label{sec:appendix_task1_prompt}

\paragraph{Chinese prompt.}
\begin{quote}\small
\begin{zh}
你是中国艺术 VQA 基准数据集的题目设计员。你只拿到"描述文本"，但你要设计在评测时由 VLM 通过图像（P1）或图像+常识（P2）回答的题目。
\end{zh}

(You are a question designer for a Chinese-art VQA benchmark. You only have access to a textual description, but you must design questions answerable by a VLM using the image (P1) or image + commonsense (P2).)

\begin{zh}
\textbf{硬性约束：}
\end{zh}

(\textbf{Hard constraints:})
\begin{itemize}\small
  \item \begin{zh}只输出 P1 或 P2。\end{zh} (Output only P1 or P2.)
  \item \begin{zh}不要生成需要依赖"标题/作者/年代/著录/历史事件细节/释文"等文本信息才能回答的题。\end{zh} (Do not generate questions requiring title/author/date/provenance/historical details/transcription.)
  \item \begin{zh}每题必须给出2到5条预期可见线索，并提供一句说明为什么不需要阅读描述文本也能回答。\end{zh} (Each question must provide 2--5 expected visual cues and a one-sentence explanation of why it is answerable without the text.)
  \item \begin{zh}只能写"预期的图像证据/可见线索"，不得引用或复述描述原句。\end{zh} (Only write expected image evidence/visual cues; do not quote the description.)
\end{itemize}

\begin{zh}
\textbf{题型覆盖要求：}生成6到10题，只允许 P1 或 P2，并尽量覆盖：
\end{zh}

(\textbf{Coverage requirement:} Generate 6--10 questions, P1/P2 only, covering:)
\begin{itemize}\small
  \item QA1 \begin{zh}主体/对象识别\end{zh} (Subject/object recognition, P1)
  \item QA2 \begin{zh}场景/活动归类\end{zh} (Scene/activity classification, P2)
  \item QA3 \begin{zh}形制/形式\end{zh} (Composition/format: scroll/album/fan, P1/P2)
  \item QA4 \begin{zh}技法线索\end{zh} (Visible technique/style cues, P1)
  \item QA5 \begin{zh}风格$\rightarrow$时期推断\end{zh} (Style$\rightarrow$period inference; only when description implies visible cues; must be P2; phrase as ``which period is more likely'' with ``unable to determine'' option)
  \item QA6 \begin{zh}可见符号/对象\end{zh} (Visible iconography/elements, P1)
\end{itemize}
\end{quote}

\subsection{Prompt for \catalogcaption}
\label{sec:appendix_task2_prompt}

\paragraph{Chinese prompt.}
\begin{quote}\small
\begin{zh}
你将收到一张中国艺术作品的图片。请用中文输出"专业鉴赏式"描述，并严格遵循以下格式与要求：
\end{zh}

(You will be given an image of a Chinese artwork. Write a professional, connoisseur-style appreciation in Chinese, strictly following the format below.)

\begin{zh}
\textbf{格式要求：}必须按顺序输出四个小节标题，且每节 1--3 句：
\end{zh}

(\textbf{Format:} Output four section headers in order, 1--3 sentences each:)
\begin{itemize}\small
  \item \begin{zh}【作品背景】\end{zh} ([Artwork Background]: Make reasonable inferences; if uncertain, state ``hard to determine from the image alone.'' Do not fabricate author/site/institution.)
  \item \begin{zh}【作品内容】\end{zh} ([Artwork Content]: Describe subject, theme, composition, key visible details.)
  \item \begin{zh}【艺术特色】\end{zh} ([Artistic Characteristics]: Discuss material, technique, color, lines, style cues from visible evidence.)
  \item \begin{zh}【综合评价】\end{zh} ([Overall Evaluation]: Provide aesthetic judgment explaining why it is compelling. Avoid generic praise.)
\end{itemize}

\begin{zh}
\textbf{约束：}若某信息无法从图像可靠判断，请明确说明不确定，不要编造。语言要求专业、具体、有依据。总长度建议不超过 1200 字。
\end{zh}

(\textbf{Constraints:} If uncertain, state so explicitly. Keep writing professional, specific, evidence-grounded. Recommended length: $\le$1200 characters.)
\end{quote}

\subsection{Prompt for \reinterpret}
\label{sec:appendix_task3_prompt}

\paragraph{Chinese prompt.}
\begin{quote}\small
\begin{zh}
你将收到一张艺术品图片。请用中文写一段"创新赏析"风格的鉴赏文本，尽量依据可见内容，不要编造无法从图像判断的具体作者、出土地点或藏馆信息；不确定就说明"仅凭图像难以确定"。
\end{zh}

(You will be given an artwork image. Write an ``innovative appreciation'' in Chinese, grounded in visible evidence. Do not invent specific facts not determinable from the image; if uncertain, state so.)

\begin{zh}
输出请严格包含以下四个小节标题（每节 1 至 3 句，信息密度高）：
\end{zh}

(Output must include four section headers, 1--3 sentences each, high information density:)
\begin{itemize}\small
  \item \begin{zh}【背景】\end{zh} ([Background]: era, use, cultural context; reasonable inference allowed)
  \item \begin{zh}【作品内容】\end{zh} ([Content]: main subject, theme, composition, key visual details)
  \item \begin{zh}【艺术特色】\end{zh} ([Artistic Characteristics]: material, craft, technique, linework, coloration, style)
  \item \begin{zh}【意义价值】\end{zh} ([Significance]: artistic value, historical/cultural significance, influence)
\end{itemize}

\begin{zh}
总长度建议不超过 \texttt{<MAX\_CHARS>} 字。
\end{zh}

(Recommended length: $\le$\texttt{<MAX\_CHARS>} characters.)
\end{quote}

\subsection{Prompt for \connoisseurpairs}
\label{sec:appendix_task4_prompt}

\paragraph{Chinese prompt.}
\begin{quote}\small
\begin{zh}
你将看到一张对比图：左侧为图像A，右侧为图像B（图上角标已标注A/B）。两幅图为同一题材作品的一真一仿。
\end{zh}

(You will see a side-by-side comparison: Image A on the left, Image B on the right. One is authentic, the other an imitation of the same subject.)

\begin{zh}
\textbf{任务：}判断【哪一张更可能是真迹】（只能选A或B）。
\end{zh}

(\textbf{Task:} Decide which is more likely authentic. Choose only A or B.)

\begin{zh}
\textbf{依据限制：}只能依据图像证据作判断，不得依赖标题、作者、年代、收藏信息等外部知识。
\end{zh}

(\textbf{Restriction:} Base decision only on visible evidence. Do not use external metadata.)

\begin{zh}
\textbf{输出要求：}
\end{zh}

(\textbf{Output:})
\begin{itemize}\small
  \item (1) \begin{zh}结论\end{zh} (Decision): A or B.
  \item (2) \begin{zh}理由\end{zh} (Reasons): 3--6 points grounded in visual cues (brushwork, ink layering, texture strokes, color, support texture, seal consistency, copying artifacts).
  \item (3) \begin{zh}不确定性\end{zh} (Uncertainty): If evidence is close, state the key uncertainty and what additional visual details would be needed.
\end{itemize}
\end{quote}

\begin{figure*}[t]
  \centering
  \includegraphics[width=0.85\linewidth]{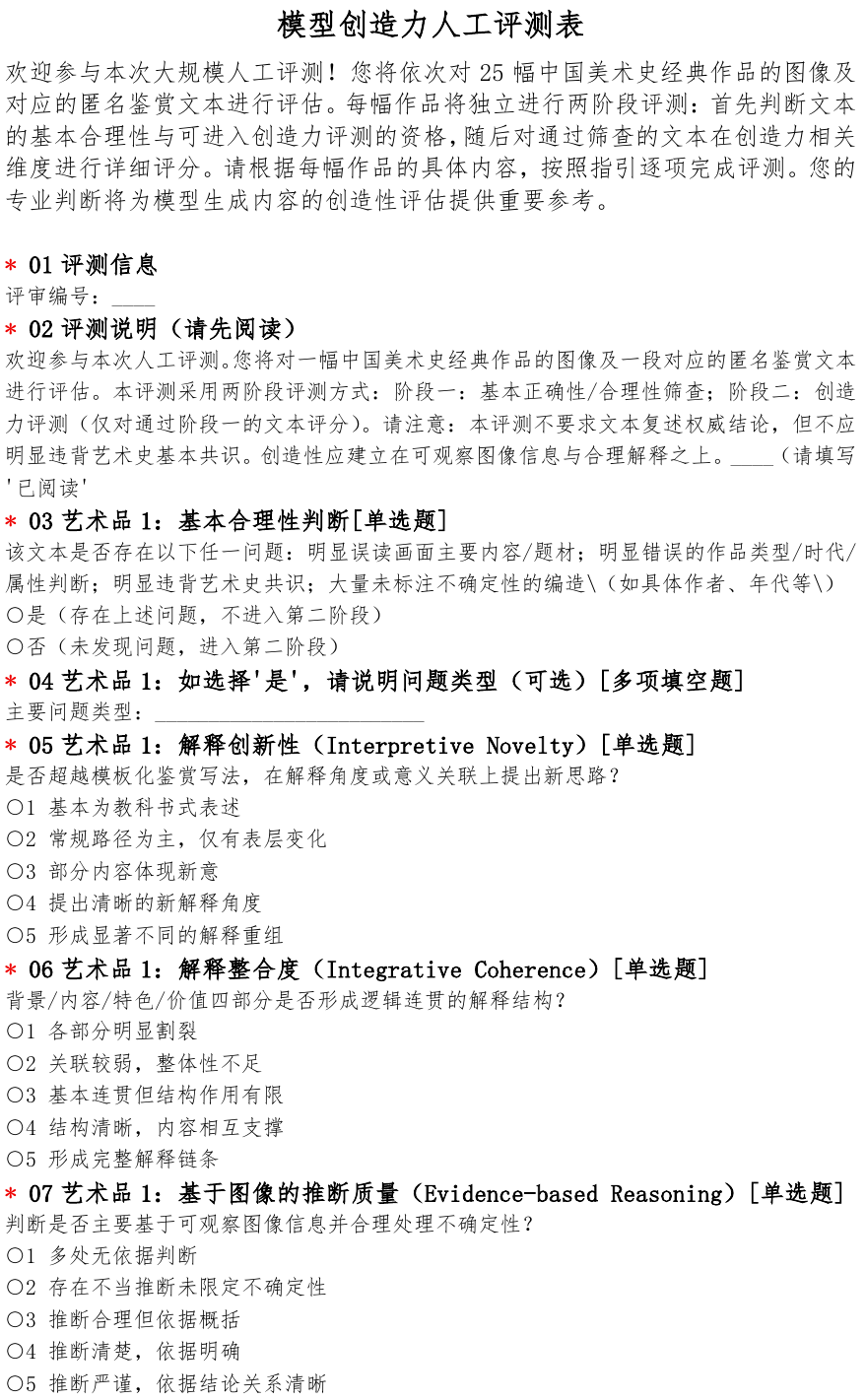}
   \caption{Survey Questionnaire for \reinterpret{}-Part1.}
  \label{fig:model_1_1}
\end{figure*}

\begin{figure*}[t]
  \centering
  \includegraphics[width=0.85\linewidth]{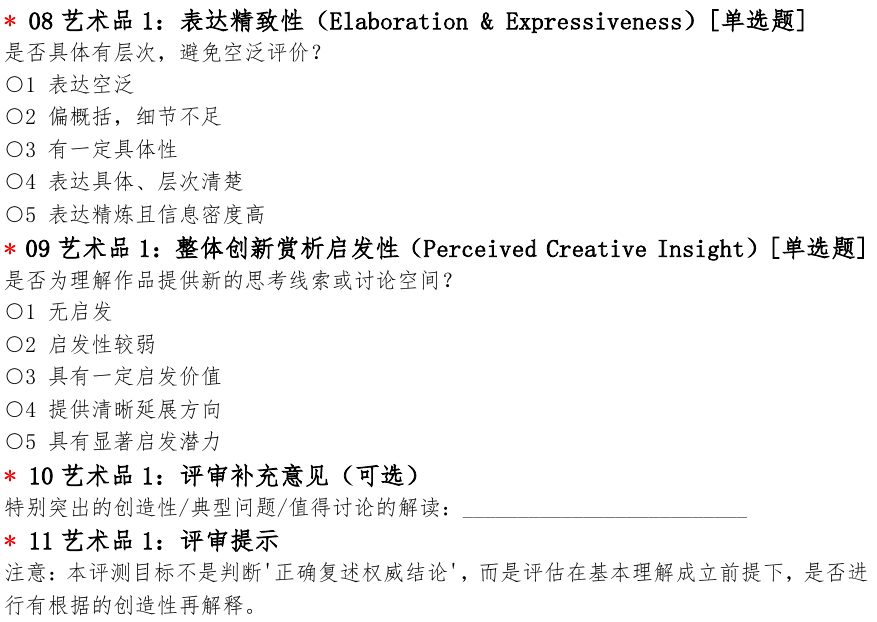}
  \caption{Survey Questionnaire for \reinterpret-Part2.}
  \label{fig:model_1_2}
\end{figure*}

\begin{figure*}[t]
  \centering
  \includegraphics[width=0.85\linewidth]{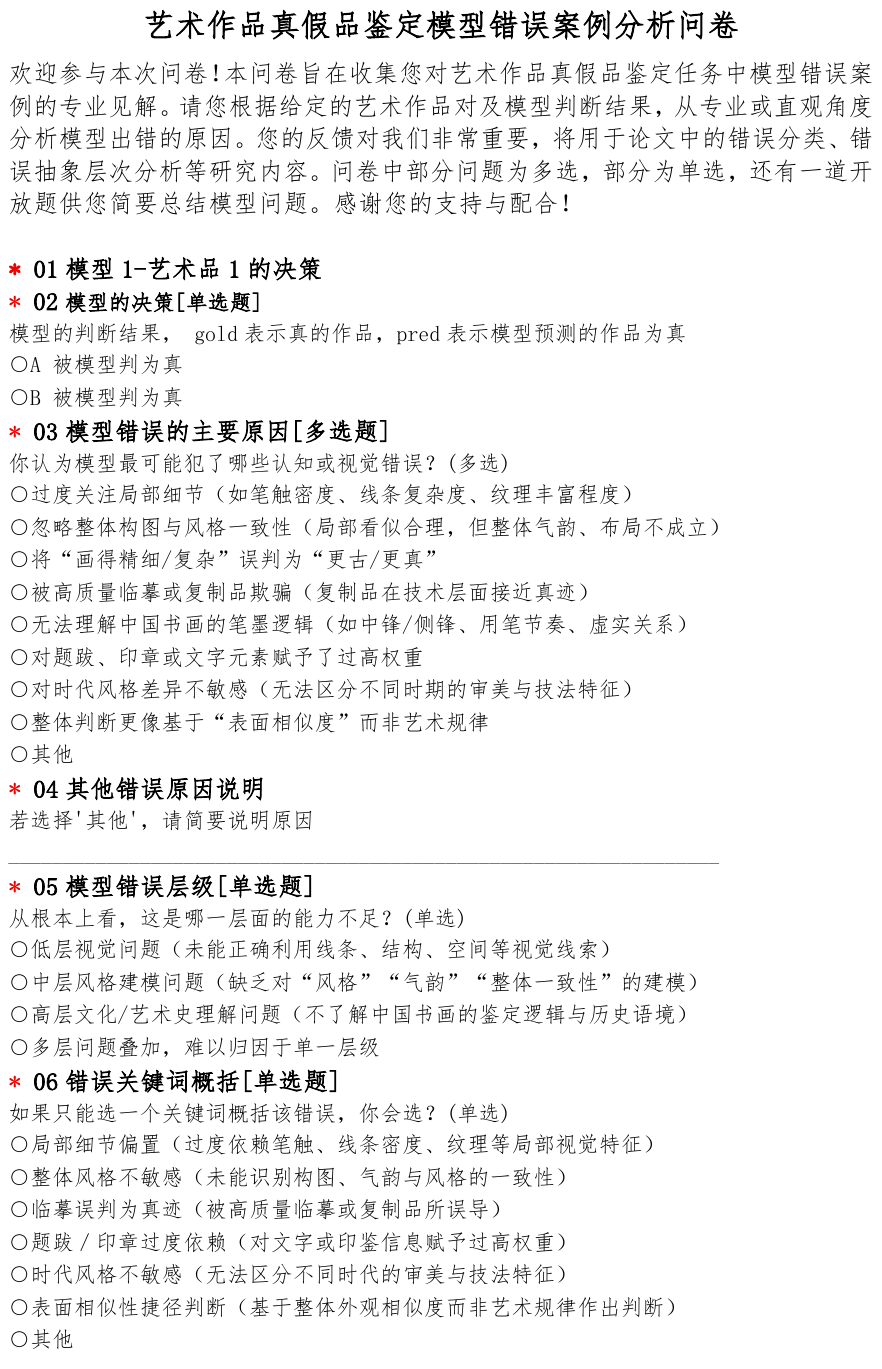}
\caption{Survey Questionnaire for \connoisseurpairs-Part1.}
  \label{fig:creative_1_1}
\end{figure*}

\begin{figure*}[t]
  \centering
  \includegraphics[width=0.85\linewidth]{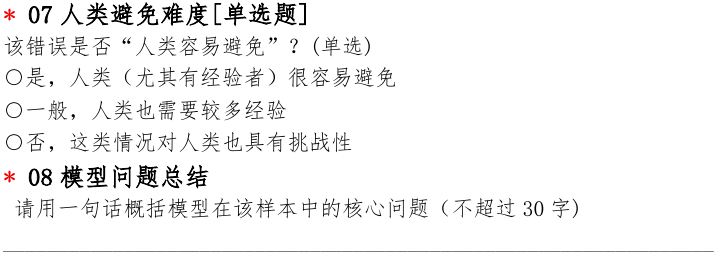}
  \caption{Survey Questionnaire for \connoisseurpairs-Part2.}
  \label{fig:creative_1_2}
\end{figure*}

\subsection{Survey instrument for \reinterpret}
\label{sec:appendix_task3_survey}

We evaluate \reinterpret with a two-stage human questionnaire designed for Chinese-art ``creative reinterpretation.'' The original survey is shown in Figure~\ref{fig:model_1_1} and Figure~\ref{fig:model_1_2}.
Raters are presented with an artwork image and an anonymous model-generated interpretation.
Stage~1 is a plausibility gate that filters outputs with any of the following issues: (i) clear misreading of the main content/theme,
(ii) clearly wrong attribution of artwork type/era/attributes, (iii) claims that contradict widely accepted art-historical commonsense,
or (iv) substantial fabrication of specific facts (e.g., author, exact era) without explicitly stating uncertainty.
If any issue is detected, the rater marks the output as \emph{fail} and the instance does not enter Stage~2.

Stage~2 rates only gate-passed outputs on five 1--5 Likert dimensions:
\textbf{D1 Interpretive Novelty}, \textbf{D2 Integrative Coherence}, \textbf{D3 Evidence-based Reasoning},
\textbf{D4 Elaboration \& Expressiveness}, and \textbf{D5 Perceived Creative Insight}.
Each dimension uses anchored options to reduce subjectivity. Briefly:
(1) Novelty ranges from textbook-like restatement to a clearly distinctive reinterpretation;
(2) Coherence ranges from fragmented structure to a complete and well-supported interpretation chain;
(3) Evidence-based Reasoning ranges from unsupported assertions to rigorous inference with clear evidence links;
(4) Elaboration ranges from vague and generic to concrete, layered, and information-dense expression;
(5) Insight ranges from no new takeaway to a strong, discussion-worthy perspective.
Raters can also provide optional comments for notable strengths or issues.

\subsection{Expert error-analysis questionnaire for \connoisseurpairs}
\label{sec:appendix_task4_survey}

For \connoisseurpairs, we further conduct an expert-facing error analysis on misclassified authentic--imitation pairs.
Annotators are shown the paired comparison (A/B) along with the model decision and the gold label (which image is authentic). The original survey is shown in Figure~\ref{fig:creative_1_1} and Figure~\ref{fig:creative_1_2}.
They first confirm the model's decision (A or B), then select all applicable error causes from a predefined checklist,
including: over-focusing on local detail richness, ignoring global composition and style consistency, mistaking ``more detailed/complex''
for ``older/more authentic,'' being misled by high-quality replicas, weak understanding of brush-and-ink logic,
over-weighting inscriptions and seals, insensitivity to era/style differences, or relying on surface similarity rather than connoisseurship cues.
Annotators can optionally describe additional reasons in free text.

To characterize the error at a higher level, annotators also assign a primary error layer
(low-level vision cues vs.\ mid-level style modeling vs.\ high-level cultural/art-historical understanding, or multi-factor),
choose a single best summary keyword for the failure, rate whether the error is easy for humans to avoid,
and provide a one-sentence diagnosis (limited to a short length) to support qualitative analysis and taxonomy building.

\subsection{LLM Configuration}
\label{sec:appendix_llm_config}

\paragraph{Access methods.}
We evaluated both open-weight and API-based VLMs.
OpenAI models were accessed via the OpenAI API.\footnote{\url{https://platform.openai.com/docs/overview}}
Gemini models were accessed via Google's Gemini API.\footnote{\url{https://ai.google.dev/gemini-api/docs}}
For Qwen/GLM families, we used the AI Ping API gateway (\url{https://www.aiping.cn/}) and selected model identifiers from its public model list endpoint.\footnote{\url{https://www.aiping.cn/docs/API/modelList}}

\paragraph{Model identifiers used in our experiments.}
OpenAI: \texttt{gpt-5-mini}, \texttt{gpt-5-nano} (evaluation); and \texttt{gpt-5.2} for LLM-assisted \curatorqa{} question generation.
According to OpenAI API pricing, the text-token prices (per 1M tokens) are:
\texttt{gpt-5-mini} input \$0.25 / cached input \$0.025 / output \$2.00;
\texttt{gpt-5-nano} input \$0.05 / cached input \$0.005 / output \$0.40;
\texttt{gpt-5.2-2025-12-11} input \$1.75 / cached input \$0.175 / output \$14.00.\footnote{\url{https://platform.openai.com/docs/pricing}}

Gemini: \texttt{gemini-2.5-flash}. The Gemini API pricing page reports (standard, per 1M tokens) \$0.30 for input text/image/video and \$2.50 for output; and (batch) \$0.15 for input and \$1.25 for output.\footnote{\url{https://ai.google.dev/pricing}}

AI Ping model IDs included:
qwen3-vl-235b-a22b-thinking, 
qwen3-vl-30b,
qwen2.5-vl-72b, 
qwen2.5-vl-32b, 
qwen2.5-vl-7b,
glm-4.5v, and glm-4.6v.

AI Ping exposes input price range and output price range fields for each model in the same model list endpoint, and we report costs following these service-provided prices.\footnote{\url{https://aiping.cn/api/v1/models}}

\paragraph{Decoding and inference hyperparameters.}
We use deterministic decoding for all models with temperature=0 and top\_p=1.
Unless otherwise noted, we generate a single response per prompt (\texttt{num\_samples}=1). And we set the max\_output=16k.

\subsection{Dynasty timeline used for era statistics}
\label{sec:appendix_dynasty_timeline}

\begin{figure*}[!t]
  \centering
  \includegraphics[width=\linewidth]{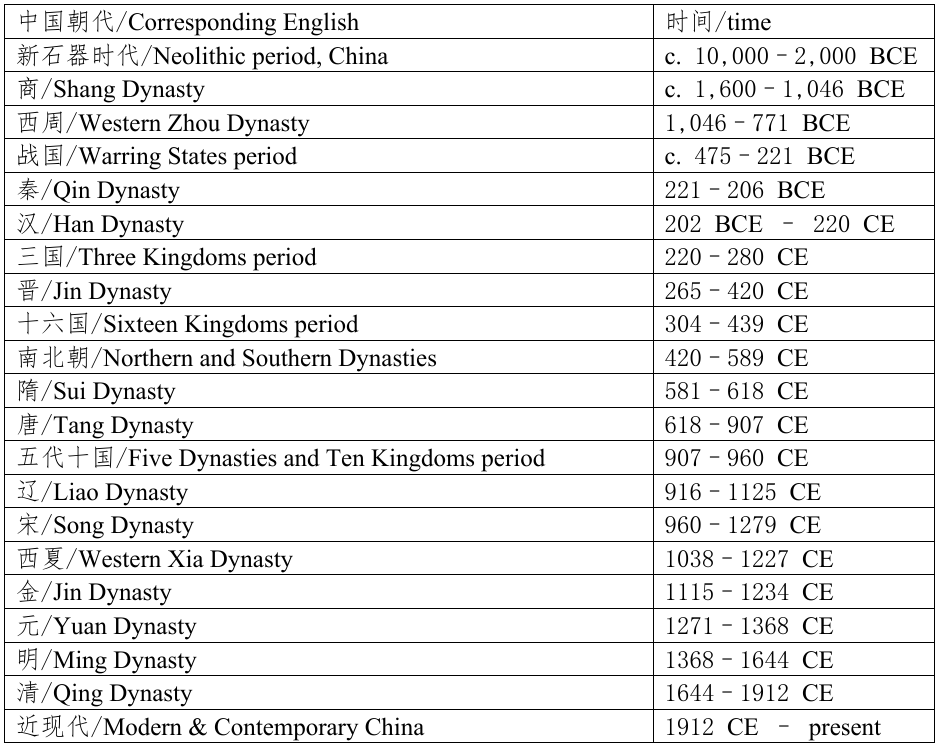}
  \caption{Chinese dynasties and historical periods considered in this work, listing the Chinese names with English equivalents and their approximate date ranges (BCE/CE).}
  \label{fig:dynasty_timeline}
\end{figure*}

To make the era distribution in Figure~\ref{fig:3} interpretable to readers unfamiliar with Chinese history, we provide in Figure~\ref{fig:dynasty_timeline} a reference timeline of the dynasties and historical periods used in this work.
We follow a widely adopted periodization in Chinese art history and use these ranges as \emph{coarse} bins when reporting era coverage.
Here, \textit{BCE} denotes ``Before Common Era'' and \textit{CE} denotes ``Common Era''; \textit{c.} (circa) indicates an approximate date range.
Some periods correspond to transitional eras or partially overlapping regimes (e.g., the Sixteen Kingdoms period), and museum records may use finer-grained attributions.
For consistency in our analysis, we map such metadata to the closest period in Figure~\ref{fig:dynasty_timeline}.

\subsection{Data stats}
\label{appendix:data-stats}
Table~\ref{tab:curatorqa-summary} reports the main statistics and distributions.

\begin{table*}[t]
\centering
\small
\begin{tabularx}{\textwidth}{@{}l l X r r@{}}
\toprule
\textbf{Group} & \textbf{Tag} & \textbf{Description} & \textbf{Numbers} & \textbf{Percentage} \\
\midrule
\multicolumn{5}{@{}l}{\textbf{Overview}} \\
Entries   & -- & Unique artwork entries & 1{,}589 & -- \\
Questions & -- & Total number of questions & 14{,}421 & -- \\
Avg. Q / entry & -- & Average questions per entry & 9.07 & -- \\
\midrule
\multicolumn{5}{@{}l}{\textbf{Difficulty}} \\
Principle & P1 & Visual-only (answerable from the image) & 10{,}920 & 75.72 \\
Principle & P2 & Vision plus art commonsense/knowledge & 3{,}501 & 24.27 \\
\midrule
\multicolumn{5}{@{}l}{\textbf{Question types}} \\
QA type & QA1 & Subject recognition (primary subject/object) & 2{,}135 & 14.80 \\
QA type & QA2 & Scene/activity classification & 1{,}932 & 13.39 \\
QA type & QA3 & Composition/format (form, layout, format) & 2{,}578 & 17.87 \\
QA type & QA4 & Visible technique/style cues & 2{,}882 & 19.98 \\
QA type & QA5 & Style$\rightarrow$period inference & 1{,}493 & 10.35 \\
QA type & QA6 & Iconography (visible symbols/elements) & 3{,}401 & 23.58 \\
\bottomrule
\end{tabularx}
\caption{Summary statistics of \curatorqa, including overview counts, difficulty principles, and question-type distribution.}
\label{tab:curatorqa-summary}
\end{table*}

\end{document}